\documentclass[preprint]{imsart}

\RequirePackage[authoryear,round,sort]{natbib}
\RequirePackage[colorlinks,citecolor=blue,urlcolor=blue]{hyperref}

\usepackage[utf8]{inputenc}
\usepackage[T1]{fontenc}
\usepackage{amsfonts,stmaryrd,amsmath,amsthm,algorithm,algorithmicx,bbm,eucal,placeins}
\usepackage{fouriernc}
\usepackage{adjustbox}
\usepackage{subcaption}
\usepackage[english]{babel}
\usepackage{tikz}
\usepackage{mathtools}
\doi{}
\pubyear{}
\volume{}
\firstpage{}
\lastpage{}
\arxiv{}

\startlocaldefs

\newtheorem{theo}{Theorem}%[section]
\newtheorem{model}{Model}
\newtheorem{prop}{Proposition}%[section]
\newtheorem{coro}{Corollary}%[section]
\newtheorem{lem}{Lemma}%[section]
\newtheorem{rem}{Remark}

\graphicspath{{fig/}}
\allowdisplaybreaks
\def\name{\textsc{PACBO}}
\def\R{\mathbb{R}}
\def\E{\mathbb{E}}
\def\ie{\emph{i.e.}}
\def\PP{\mathcal{P}}
\def\K{\mathcal{K}}
\def\CC{\mathcal{C}}
\def\cl{K_{t}}
\def\bc{\mathbf{c}}

\def\ttc{\mathfrak{c}}
\def\Rkn{\mathbb{R}^{dk^{(n)}}}

\def\expcum{\sum_{t=1}^T\mathbb{E}_{(\hat{\rho}_{1},\hat{\rho}_{2},\dots,\hat{\rho}_{t})}\ell(\hat{\bc}_{t},x_{t})}
\def\rdexpcum{\sum_{t=1}^T\mathbb{E}_{(\hat{\rho}_{1},\hat{\rho}_{2},\dots,\hat{\rho}_{t})}\ell(\hat{\bc}_{t},X_{t})}
\def\infpart{\inf_{\bc \in \CC(s,R)}\sum_{t=1 }^T\ell(\bc,X_t)}

\DeclarePairedDelimiter\ceil{\lceil}{\rceil}

\endlocaldefs

\begin{document}

\begin{frontmatter}

% "Title of the Paper"
\title{A Quasi-Bayesian Perspective to Online Clustering}
% \thankstext{t1}{This is an original survey paper}
\runtitle{A Quasi-Bayesian Perspective to Online Clustering}

% indicate corresponding author with \corref{}
\author{\fnms{Le} \snm{Li}\ead[label=e3]{le@iadvize.com}}
\address{Université d'Angers \& iAdvize\\ \printead{e3}}

\author{\fnms{Benjamin} \snm{Guedj}\thanksref{t1}\ead[label=e1]{benjamin.guedj@inria.fr}\ead[label=e2,url]{https://bguedj.github.io}}
\thankstext{t1}{Corresponding author} 
\address{Inria\\ Modal project-team, Lille - Nord Europe research center\\ France\\ \printead{e1}\\ \printead{e2}}
\medskip
\and

\author{\fnms{Sébastien} \snm{Loustau}\ead[label=e4,url]{http://www.math.univ-angers.fr/~loustau/}\ead[label=e5,url]{http://www.lumenai.fr}\ead[label=e6]{artfact64@gmail.com}}
\address{Lumen AI\\
% \printead{e4}\\
\printead{e5}
% \\
% \printead{e6}
}

\runauthor{Li, Guedj and Loustau}

\begin{abstract}
\noindent When faced with high frequency streams of data, clustering raises theoretical and algorithmic pitfalls. We introduce a new and adaptive online clustering algorithm relying on a quasi-Bayesian approach, with a dynamic (\ie, time-dependent) estimation of the (unknown and changing) number of clusters. We prove that our approach is supported by minimax regret bounds. We also provide an RJMCMC-flavored implementation (called \name, see \href{https://cran.r-project.org/web/packages/PACBO/index.html}{https://cran.r-project.org/web/packages/PACBO/index.html}) for which we give a convergence guarantee. Finally, numerical experiments illustrate the potential of our procedure.
\end{abstract}



% history:
%\received{\smonth{12} \syear{2017}}

\tableofcontents

\end{frontmatter}

\usetikzlibrary{arrows}

% Heading arguments are {volume}{year}{pages}{submitted}{published}{author-full-names}

%%%% Librairies TIKZ

% \begin{abstract}%   <- trailing '%' for backward compatibility of .sty file

% \end{abstract}

% \begin{keywords}

% \end{keywords}

%\tableofcontents

\section{Introduction}\label{sec:intro}
Online learning has been extensively studied these last decades in game
theory and statistics \citep[see][and references therein]{CBL2006}. The problem
can be described as a sequential game: a blackbox reveals at each time $t$ some $z_t\in\mathcal{Z}$. Then, the forecaster predicts the next value based on
the past observations and possibly other available information. In the present work we will consider the scenario in which the sequence $(z_{t})$ is not assumed to be a realization of some stochastic process. One of the well known problem in online learning that happened to attract a lot of interest is prediction with expert advice. In this setting, the forecaster has access to a set $\{f_{e,t}\in\mathcal{D} : e\in \mathcal{E}\}$ of experts' predictions, where $f_{e,t}$ is the prediction of expert $e$ at time $t$, $\mathcal{D}$ is a decision space which is assumed to be a convex subset of vector space and $\mathcal{E}$ is a finite set of experts (such as deterministic physical models, or stochastic decisions). Predictions made by the forecaster and experts are assessed with a loss function $\ell:\mathcal{D}\times \mathcal{Z}\longrightarrow \R_{+}$. The goal is to build a sequence $\hat{z}_1,\dots,\hat{z}_T$ (denoted by $(\hat{z}_{t})_{1:T}$ in the sequel) of predictions which are nearly as good as the best expert's predictions in the first $T$ time rounds, \ie, satisfying uniformly over any sequence $(z_{t})$ the following regret bound

\begin{equation*}
\sum_{t=1}^{T}\ell\left(\hat{z}_{t},z_{t}\right) - \min_{e \in \mathcal{E}}\left\{\sum_{t=1}^{T}\ell\left(f_{e,t},z_{t}\right)\right\} \leq \Delta_{T}(\mathcal{E}),
\end{equation*}
where $\Delta_{T}(\mathcal{E})$ is a remainder term. This term should be as small as possible and in particular sublinear in $T$. When $\mathcal{E}$ is finite, and the loss is bounded in $[0,1]$ and convex in its first argument, an optimal $\Delta_{T}(\mathcal{E})=\sqrt{(T/2)\log \vert \mathcal{E}\vert}$ is given by Theorem 2.2 of \cite{CBL2006}. The optimal forecaster is then obtained by forming the exponentially weighted average of all experts. For similar results, we refer the reader to \cite{LW1994} and \cite{CFH1997}.
\medskip

Online learning techniques have also been applied to the regression framework. In particular, sequential ridge regression has been studied by \cite{Vov2001}.
For any $t=1,\dots,T$, we now assume that $z_{t}=(x_{t},y_{t})\in \R^{d}\times\R$. At each time $t$, the forecaster gives a prediction $\hat{y}_{t}$ of $y_{t}$, using only newly revealed side information $x_{t}$ and past observations $(x_{s},y_{s})_{1:(t-1)}$. Let $\langle \cdot,\cdot \rangle$ denote the scalar product in $\R^{d}$. A possible goal is to build a forecaster whose performance is nearly as good as the best linear forecaster $f_{\theta}\colon x\mapsto \langle \theta,x\rangle$, \ie, such that uniformly over all sequences $(x_{t},y_{t})_{1:T}$,
\begin{equation}\label{eq:regret bound for online linear regression}
\sum_{t=1}^{T}\ell\left(\hat{y_{t}},y_{t}\right)-\inf_{\theta\in\R^{d}}\left\{\sum_{t=1}^{T}\ell\left(\langle \theta,x_{t}\rangle,y_{t}\right)\right\}\leq \Delta_{T}(d),
\end{equation}
where $\Delta_{T}(d)$ is a remainder term. This setting has been addressed by numerous contributions to the literature. In particular, \cite{AW2001} and \cite{Vov2001} each provide an algorithm close to the ridge regression with a remainder term $\Delta_{T}(d)=\mathcal{O}(d\log T)$. Other authors have investigated the Gradient-Descent algorithm \citep{CLW1996,KW1997} and the Exponentiated Gradient Forecasters \citep{KW1997,CB1999}. \cite{Ger2011} extended the linear form $\langle u,x_{t}\rangle$  in \eqref{eq:regret bound for online linear regression} to $\langle u,\mathbf{\varphi}(x_{t})\rangle$ \ $=\sum_{j=1}^{d}u_{j}\varphi_{j}(x_{t})$, where $\mathbf{\varphi}=(\varphi_{1},\dots,\varphi_{d})$ is a dictionary of base forecasters. In the so-called high dimensional setting ($d\gg T$), a sparsity regret bound with a remainder term $\Delta_T(d)$ growing logarithmically with $d$ and $T$ is proved by \citet[][Proposition 3.1]{Ger2011}.
\medskip

The purpose of the present work is to generalize the aforecited framework to the clustering problem, which has attracted attention from the machine learning and streaming communities. As an example, \cite{GMM2003}, \cite{BF2008} and \cite{LSS2015} study the so-called data streaming clustering problem. It amounts to clustering online data to a fixed number of groups in a single pass, or a small number of passes, while using little memory. From a machine learning perspective, \cite{CM2012} aggregate online clustering algorithms, with a fixed number $K$ of centers. 
%\medskip
The present paper investigates a more general setting since we aim to perform online clustering with a varying number $K_t$ of centers. To the best of our knowledge, this is the first attempt of the sort in the literature. Let us stress that our approach only requires an upper bound $p$ to $K_t$, which can be either a constant or an increasing function of the time horizon $T$. 
\medskip

Our approach strongly relies on a quasi-Bayesian methodology. The use of quasi-Bayesian estimators is especially advocated by the PAC-Bayesian theory which originates in the machine learning community in the late 1990s, in the seminal works of \cite{STW1997} and \citet{McA1998,McA1999} \citep[see also][]{See2002, See2003}. In the statistical learning community, the PAC-Bayesian approach has been extensively developed by \citet{Cat2004, Cat2007}, \citet{Aud2004b} and \citet{Alq2006}, and later on adapted to the high dimensional setting \citet{DT2007,DT2008}, \cite{AL2011}, \citet{AB2013}, \citet{GA2013}, \cite{GR2015} and \citet{AG2016}. In a parallel effort, the online learning community has contributed to the PAC-Bayesian theory in the online regression setting \citep{KW1999}. \cite{Aud2009} and \cite{Ger2011} have been the first attempts to merge both lines of research. Note that our approach is \emph{quasi-Bayesian} rather than PAC-Bayesian, since we derive regret bounds (on quasi-Bayesian predictors) instead of PAC oracle inequalities.
\medskip

Our main contribution is to generalize algorithms suited for supervised learning to the unsupervised setting. Our online clustering algorithm is adaptive in the sense that it does not require the knowledge of the time horizon $T$ to be used and studied. The regret bounds that we obtain have a remainder term of magnitude $\sqrt{T\log T}$ and we prove that they are asymptotically minimax optimal.
\medskip

The quasi-posterior which we derive is a complex distribution and direct sampling is not available. In Bayesian and quasi-Bayesian frameworks, the use of Markov Chain Monte Carlo (MCMC) algorithms is a popular way to compute estimates from posterior or quasi-posterior distributions. We refer to the comprehensive monograph \cite{RG2004} for an introduction to MCMC methods. For its ability to cope with transdimensional moves, we focus on the Reversible Jump MCMC algorithm from \cite{Gre1995}, coupled with ideas from the Subspace Carlin and Chib algorithm proposed by \cite{DFN2002} and \cite{PD2012}. MCMC procedures for quasi-Bayesian predictors were firstly considered by \cite{Cat2004} and \cite{dalalyan2012sparse}. \cite{AB2013}, \cite{GA2013} and \cite{GR2015} are the first to have investigated the RJMCMC and Subspace Carlin and Chib techniques and we show in the present paper that this scheme is well suited to the clustering problem.
\medskip

The paper is organised as follows. \hyperref[sec:notation]{Section~\ref*{sec:notation}} introduces our notation and our online clustering procedure.
\hyperref[sec:online clustering method]{Section~\ref*{sec:online clustering method}} contains our mathematical claims, consisting in regret bounds for our online clustering algorithm. Remainder terms which are sublinear in $T$ are obtained for a model selection-flavored prior. We also prove that these remainder terms are minimax optimal. We then discuss in \hyperref[sec:algo]{Section~\ref*{sec:algo}} the practical implementation of our method, which relies on an adaptation of the RJMCMC algorithm to our setting. In particular, we prove its convergence towards the target quasi-posterior. The performance of the resulting algorithm, called \name, is evaluated on synthetic data. For the sake of clarity, proofs are postponed to \hyperref[sec:proofs]{Section~\ref*{sec:proofs}}. Finally, Appendix \ref{app} contains an extension of our work to the case of a multivariate Student prior along with additional numerical experiments.

\section{A quasi-Bayesian perspective to online clustering}
\label{sec:notation}
Let $(x_{t})_{1:T}$ be a sequence of data, where $x_t\in\R^d$. Our goal is to learn a time-dependent parameter $\cl$ and a partition of the observed points into $\cl$ cells, for any $t=1,\dots,T$. To this aim, the output of our algorithm at time $t$ is a vector $\hat{\bc}_{t}=(\hat{c}_{t,1},\hat{c}_{t,2},\dots,\hat{c}_{t,\cl})$ of $\cl$ centers in $\R^{d\cl}$, depending on the past information $(x_{s})_{1:(t-1)}$ and $(\hat{\bc}_{s})_{1:(t-1)}$.  A partition is then created by assigning any point in $\R^d$ to its closest center. When $x_{t}$ is newly revealed, the instantaneous loss is computed as
\begin{equation}\label{eq:dynamic number of centers}
\ell(\hat{\bc}_{t},x_{t})=\min_{1\leq k\leq \cl}\vert\hat{c}_{t,k}-x_{t}\vert^{2}_{2},
\end{equation}
where $\vert\cdot\vert_{2}$ is the $\ell_{2}$-norm in $\R^d$. In what follows, we investigate regret bounds for cumulative losses. Given a measurable space $\Theta$ (embedded with its Borel $\sigma$-algebra), we let $\PP(\Theta)$ denote the set of probability distributions on $\Theta$, and for some reference measure $\nu$, we let $\PP_\nu(\Theta)$ be the set of probability distributions absolutely continuous with respect to $\nu$. For any probability distributions $\rho,\pi \in \PP(\Theta),$ the Kullback-Leibler  divergence $\K(\rho,\pi)$ is defined as
\begin{equation*}
\K(\rho,\pi)=
\begin{cases}
\int_{\Theta} \log \left(\frac{\mathrm{d}\rho}{\mathrm{d}\pi}\right)\mathrm{d}\rho &\quad
\text{when } \rho \in \PP_\pi(\Theta),
 \\
 +\infty &\quad \text{otherwise.}
\end{cases}
\end{equation*}

Note that for any bounded measurable function $h\colon\Theta \to \R$ and any probability distribution $\rho \in \PP(\Theta)$ such that $\K(\rho,\pi)< +\infty$,
\begin{equation}\label{eq:duality formula}
-\log \int_{\Theta} \exp(-h)\mathrm{d}\pi=\inf_{\rho \in \PP(\Theta)}\left\{\int_{\Theta}h\mathrm{d}\rho+\K(\rho,\pi)\right\}.
\end{equation}
This result, which may be found in \cite{C75} and \citet[][Equation 5.2.1]{Cat2004}, is critical to our scheme of proofs. Further, the infimum is achieved at the so-called Gibbs quasi-posterior $\hat{\rho}$, defined by
\begin{equation*}
\mathrm{d}\hat{\rho}=\frac{\exp(-h)}{\int \exp(-h)\mathrm{d}\pi}\mathrm{d}\pi.
\end{equation*}

We now introduce the notation to our online clustering setting. Let $\mathcal{C}=\cup_{k=1}^{p}\R^{dk}$ for some integer $p\geq 1$. We denote by $q$ a discrete probability distribution on the set $\llbracket 1,p\rrbracket:=\{1,\dots, p\}$. For any $k\in \llbracket 1,p\rrbracket$, let $\pi_{k}$ denote a probability distribution on $\R^{dk}$.
For any vector of cluster centers $\bc \in \mathcal{C}$,  we define $\pi(\bc)$ as
\begin{equation}\label{eq:prior distribution}
\pi(\bc)=\sum_{k\in \llbracket 1,p\rrbracket}q(k)\mathbbm{1}_{\left\{\bc \in \R^{dk}\right\}}\pi_{k}(\bc).
\end{equation}
Note that \eqref{eq:prior distribution} may be seen as a distribution over the set of Voronoi partitions of $\R^d$: any $\bc \in \mathcal{C}$ corresponds to a Voronoi partition of $\R^{d}$ with at most $p$ cells. In the sequel, we denote by $\bc\in\mathcal{C}$ either a vector of centers or its associated Voronoi partition of $\R^d$ if no confusion arises, and we denote by $\pi\in\mathcal{P}(\mathcal{C})$ a prior over $\mathcal{C}$.  Let $\lambda>0$ be some (inverse temperature) parameter.
At each time $t$, we observe $x_t$ and a random partition $\hat{\bc}_{t+1}\in \mathcal{C}$ is sampled from the Gibbs quasi-posterior
\begin{equation}
\label{eq:target function}
\mathrm{d}\hat{\rho}_{t+1}(\bc)\propto\exp\big(-\lambda S_{t}(\bc)\big)\mathrm{d}\pi(\bc).
\end{equation}
This quasi-posterior distribution will allow us to sample partitions with respect to the prior $\pi$ defined in \eqref{eq:prior distribution} and bent to fit past observations through the following cumulative loss 
\begin{equation*}
S_{t}(\bc)=S_{t-1}(\bc)+\ell(\bc,x_{t})+\frac{\lambda}{2}\big(\ell(\bc,x_{t})-\ell(\hat{\bc}_{t},x_{t})\big)^2,
\end{equation*}
where the latter one is a variance term. It is essential to make the \emph{online variance inequality} hold true for general loss $\ell$ with quasi-posterior distribution, \emph{i.e.,} no constraint such as the convexity or boundedness is imposed on $\ell$  \citep[as discussed in][Section 4.2]{Aud2009}. $S_t(\bc)$ consists in the cumulative loss of $\bc$ in the first $t$ rounds and a term that controls the variance of the next prediction. Note that since $(x_t)_{1:T}$ is deterministic, no likelihood is attached to our approach, hence the terms "quasi-posterior" for $\hat{\rho}_{t+1}$ and "quasi-Bayesian" for our global method. The resulting estimate is a realization of $\hat{\rho}_{t+1}$ with a random number $\cl$ of cells. This scheme is described in \autoref{proc:procedure 1}. Note that this algorithm is an instantiation of Audibert's online SeqRand algorithm \citep[][Section 4]{Aud2009} to the special case of the loss defined in \eqref{eq:dynamic number of centers}. However SeqRand does not account for adaptive rates $\lambda=\lambda_t$, as discussed in the next section.

\begin{algorithm}[h]
\caption{The quasi-Bayesian online clustering algorithm}
\label{proc:procedure 1}
\begin{algorithmic}[1]
\\\textbf{Input parameters}: $p>0,\pi\in\mathcal{P}(\mathcal{C})$, $\lambda>0$ and $S_{0}\equiv 0$
\\\textbf{Initialization}: Draw $\hat{\bc}_{1} \sim \pi=\hat{\rho}_1$
\State \textbf{For $t\in\llbracket 1,T\rrbracket$}
\State\hspace{\algorithmicindent} Get the data $x_{t}$
\State \hspace{\algorithmicindent} Draw $\hat{\bc}_{t+1} \sim \hat{\rho}_{t+1}(\bc)$ where $\mathrm{d}\hat{\rho}_{t+1}(\bc)\propto\exp\big(-\lambda S_{t}(\bc)\big)\mathrm{d}\pi(\bc)$,
and
\begin{equation*}
S_{t}(\bc)=S_{t-1}(\bc)+\ell(\bc,x_{t})+\frac{\lambda}{2}\big(\ell(\bc,x_{t})-\ell(\hat{\bc}_{t},x_{t})\big)^2.
\end{equation*}
\State \textbf{End for}
\end{algorithmic}
\end{algorithm}

\section{Minimax regret bounds}\label{sec:online clustering method}

Let  $\mathbb{E}_{\bc\sim\nu}$ stands for the expectation with respect to the distribution $\nu$ of $\bc$ (abbreviated as $\E_{\nu}$ where no confusion is possible). We start with the following pivotal result.
\begin{prop}\label{thm:theorem 1}
For any sequence $(x_{t})_{1:T} \in \R^{dT}$, for any prior distribution $\pi\in\mathcal{P}(\mathcal{C})$ and any $\lambda>0$, the procedure described in  \autoref{proc:procedure 1} satisfies
\begin{align*}
\sum_{t=1}^T\mathbb{E}
_{(\hat{\rho}_{1},\hat{\rho}_{2},\dots,\hat{\rho}_{t})}\ell(\hat{\bc}_{t},x_{t})\leq &\inf_{\rho \in \PP_\pi(\mathcal{C})}\left\{\mathbb{E}_{\bc \sim \rho}\left[\sum_{t=1}^{T}\ell(\bc,x_{t})\right]+\frac{\K(\rho,\pi)}{\lambda}\right.\\
&\left.+\frac{\lambda}{2}\mathbb{E}_{(\hat{\rho}_{1},\dots,\hat{\rho}_{T})}\mathbb{E}_{\bc \sim \rho}\sum_{t=1}^{T}[\ell(\bc,x_{t})-\ell(\hat{\bc}_{t},x_{t})]^2\right\}.
\end{align*}
\end{prop}
\autoref{thm:theorem 1} is a straightforward consequence of \citet[][Theorem 4.6]{Aud2009} applied to the loss function defined in \eqref{eq:dynamic number of centers}, the partitions $\mathcal{C}$, and any prior $\pi\in\mathcal{P}(\mathcal{C})$.
%\medskip

\subsection{Preliminary regret bounds}\label{sec:preliminary}

In the following, we instantiate the regret bound introduced in \autoref{thm:theorem 1}. Distribution $q$ in \eqref{eq:prior distribution} is chosen as the following discrete distribution on the set $\llbracket 1,p\rrbracket$
\begin{equation}\label{eq:prior on finite set} 
q(k)=\frac{\exp(-\eta k)}{\sum_{i=1}^{p}\exp(-\eta i)},\quad \eta\geq 0.
\end{equation}
When $\eta>0$, the larger the number of cells $k$, the smaller the probability mass.
Further, $\pi_k$ in \eqref{eq:prior distribution} is chosen as a product of $k$ independent uniform distributions on $\ell_2$-balls in $\mathbb{R}^{d}$:
\begin{equation}\label{eq:uniform prior}
\mathrm{d}\pi_{k}(\bc,R)=\left(\frac{\Gamma\left(\frac{d}{2}+1\right)}{\pi^{\frac{d}{2}}}\right)^{k}\frac{1}{(2R)^{dk}}\left\{\prod_{j=1}^{k}\mathbbm{1}_{\left\{\mathit{B}_{d}(2R)\right\}}(c_{j})\right\}\mathrm{d}\bc,
\end{equation}
where $R>0$, $\Gamma$ is the Gamma function and 
\begin{equation}\label{ball}
B_{d}(r)=\left\{x\in \R^{d},\, |x|_{2}\leq r\right\}
\end{equation}
is an $\ell_{2}$-ball in $\R^{d}$, centered in $0\in \R^{d}$ with radius $r>0$. Finally, for any $k \in \llbracket1,p\rrbracket$ and any $R>0$, let
\begin{equation*}
\mathcal{C}(k,R)=\left\{\bc=(c_{j})_{j=1,\dots,k}\in\R^{dk},\textrm{ such that } |c_{j}|_{2}\leq R\quad \forall j\right\}.
\end{equation*}

\begin{coro}\label{cor:corollary 1}
For any sequence $(x_{t})_{1:T}\in \R^{dT}$ and any $p\geq1$, consider $\pi$ defined by \eqref{eq:prior distribution}, \eqref{eq:prior on finite set} and \eqref{eq:uniform prior} with $\eta\geq 0$ and $R\geq \max_{t=1,\dots,T}|x_{t}|_{2}$. If $\lambda \geq (d+2)/(2TR^{2})$, the procedure described in \autoref{proc:procedure 1} satisfies
\begin{align*}
\sum_{t=1}^T\mathbb{E}_{(\hat{\rho}_{1},\hat{\rho}_{2},\dots,\hat{\rho}_{t})}\ell(\hat{\bc}_{t},x_{t})
\leq\inf_{k\in \llbracket 1,p\rrbracket}\left\{\inf_{\bc\in\mathcal{C}(k,R)}\sum_{t=1}^{T}\ell(\bc,x_{t})+\frac{dk}{2\lambda}\log \left(\frac{8R^{2}\lambda T}{d+2}\right)+\frac{\eta }{\lambda}k\right\}\\
+\left( \frac{\log p}{\lambda}+\frac{d}{2\lambda}+\frac{81\lambda TR^{4}}{2}\right),
\end{align*}
%where $C_{1}=(2R+\max_{t=1,\dots,T}|x_{t}|_{2})^{2}$.
\end{coro}
Note that $\inf_{\bc\in\mathcal{C}(k,R)}\sum_{t=1}^{T}\ell(\bc,x_{t})$ is a non-increasing function of the number $k$ of cells while the penalty is linearly increasing with $k$. Small values for $\lambda$ (or equivalently, large values for $R$) lead to small values for $k$. The additional term induced by the complexity of $\mathcal{C}=\bigcup_{k=1,\dots,p}\R^{dk}$ is $\log p$. A reasonable choice of $\lambda$ would be such that $d/\lambda \log(\lambda TR^2/d+2)$ and $\lambda TR^4$ are of the same order in $T$. The calibration $\lambda=(d+2)\sqrt{\log T}/(2\sqrt{T}R^{2})$ yields a sublinear remainder term in the following corollary.

\begin{coro}\label{cor:corollary 2}
 Under the previous notation with $\lambda=(d+2)\sqrt{\log T}/2\sqrt{T}R^{2}$, $R\geq \max_{t=1,\dots,T}|x_{t}|_{2}$ and $T>2$, the procedure described in \autoref{proc:procedure 1} satisfies
\begin{multline}\label{eq:reg}
\sum_{t=1}^T\mathbb{E}_{(\hat{\rho}_{1},\hat{\rho}_{2},\dots,\hat{\rho}_{t})}\ell(\hat{\bc}_{t},x_{t})\leq\inf_{k\in \llbracket 1,p\rrbracket}\left\{\inf_{\bc\in\mathcal{C}(k,R)}\sum_{t=1}^{T}\ell(\bc,x_{t})+\frac{2(d+\eta)R^2}{d+2}k\sqrt{T\log T} \right\} \\ +\left(\frac{2R^{2}\log p}{d+2}+\frac{{d}R^{2}}{d+2}+\frac{81(d+2)R^{2}}{4}\right)\sqrt{T\log T}.
\end{multline}
\end{coro}

\begin{rem}
 If we assume $T$ and $R$ are constants, the reason that $\lambda$ is chosen to be of order of magnitude of $d$ here, rather than of $\sqrt{d}$, is to guarantee that it satisfies the condition $\lambda \geq (d+2)/2TR^2$ in \autoref{cor:corollary 1}. However, if $T$ is sufficiently large, \emph{e.g.,} $T \geq (d+2)^{2}/d$, then the choice $\lambda = \sqrt{d\log T}/2\sqrt{T}R^{2}$ will satisfy the condition and will make the right hand side of the above inequality grow linearly in $\sqrt{d}$ while keeping the order of magnitude for $T$ and $R$. 
\end{rem}

Let us assume that the sequence $x_1, \dots, x_{T}$ is generated from a distribution with $k^{\star}\in\llbracket 1,p\rrbracket$ clusters. We then define the expected cumulative loss (ECL) and oracle cumulative loss (OCL) as
\begin{align*}
\text{ECL} &= \expcum , \\
\text{OCL} &= \underset{\bc\in \mathcal{C}(k^{\star},R)}{\inf}\sum_{t=1}^{T}\ell(\bc,x_{t}).
\end{align*}
Then \autoref{cor:corollary 2} yields
\begin{equation}\label{eq:interpretation of corollary 3.2}
\expcum -\underset{\bc\in \mathcal{C}(k^{\star},R)}{\inf}\sum_{t=1}^{T}\ell(\bc,x_{t})\leq J k^{\star}\sqrt{T \log T},
\end{equation}
where $J$ is a constant depending on $d$, $R$ and $\log p$. In \eqref{eq:interpretation of corollary 3.2} the regret of our randomized procedure, defined as the difference between ECL and OCL is sublinear in $T$. However, whenever $k^{\star}>p$, we can deduce from \autoref{cor:corollary 2} that 

\begin{align*}
\sum_{t=1}^T\mathbb{E}_{(\hat{\rho}_{1},\hat{\rho}_{2},\dots,\hat{\rho}_{t})}\ell(\hat{\bc}_{t},x_{t})- \inf_{\bc \in \mathcal{C}(k^{\star}, R)}\sum_{t=1}^{T}\ell(\bc, x_{t}) &\leq \inf_{k\in \llbracket 1,p\rrbracket}\left\{\inf_{\bc\in\mathcal{C}(k,R)}\sum_{t=1}^{T}\ell(\bc,x_{t}) - \inf_{\bc \in \mathcal{C}(k^{\star}, R)}\sum_{t=1}^{T}\ell(\bc, x_{t}) \right. \\ 
& \left. +\frac{2(d+\eta)R^{2}}{d+2}k\sqrt{T\log T} \right\} + \\
& \left(\frac{R^{2}\left(2\log p + d\right)}{d+2}+\frac{81(d+2)R^{2}}{4}\right)\sqrt{T\log T},
\end{align*}
where $\inf_{\bc \in \mathcal{C}(k^{\star}, R)}\sum_{t=1}^{T}\ell(\bc, x_{t})$ is the oracle cumulative loss (\emph{i.e.,} OCL) with $k^{\star}$ clusters.
\medskip

If there exists a $k \in \llbracket 1,p\rrbracket$ such that $\inf_{\bc\in\mathcal{C}(k,R)}\sum_{t=1}^{T}\ell(\bc,x_{t})$ is close to OCL, then our ECL is also close to OCL up to a term of order $k\sqrt{T\log T}$. However, if no such $k$ exists, then the term $\frac{2(d+\eta)R^{2}}{d+2}k\sqrt{T\log T}$ starts to dominate, hence the quality of bound is deteriorated.

\medskip

% \todo{}
% Finally, note that the dependency in $k$ in the right-hand side of \eqref{eq:reg} may be improved by choosing $\lambda=\mathcal{O}\left(\sqrt{\frac{dp}{T}} \right)$ and assuming $p=\mathcal{O}\left( \log^2 T\right)$. This allows to achieve the optimal dependency in $\sqrt{k}$ instead of $k$ in \eqref{eq:reg} and \eqref{eq:interpretation of corollary 3.2}, \emph{i.e.},
% \begin{equation*}
% \expcum -\underset{\bc\in \mathcal{C}(k^{\star},R)}{\inf}\sum_{t=1}^{T}\ell(\bc,x_{t})\leq J \sqrt{k^{\star}T\log T}.
% \end{equation*}
% However the assumption $p=\mathcal{O}\left( \log^2 T\right)$ may appear unrealistic in the online clustering setting, as $p$ may grow with $T$ at a faster rate than $\log^2 T$. The dependency in $k$ in \eqref{eq:interpretation of corollary 3.2} is the price to pay for a general framework.

% \todo{}

Finally, note that the dependency in $k$ inside the braces on the right-hand side of \eqref{eq:reg} may be improved by choosing $\lambda = (d+2)\sqrt{p\log T}/2\sqrt{T}R^{2}$ in \autoref{cor:corollary 2}. This allows to achieve the optimal rate $\sqrt{k}$ instead of $k$, since $k/\sqrt{p} \leq \sqrt{k}$ for any $k \in \llbracket 1,p\rrbracket$. However, this makes the last term in \autoref{cor:corollary 2} of order of $\sqrt{pT\log T}$. Note that the effort to make the regret bound grow in $\sqrt{k}$, rather than $\sqrt{p}$ for $k \in \llbracket 1, p\rrbracket$ may be achieved by using a similar strategy to the one of \cite{Win2017}, which introduces a recursive aggregation procedure with distinct learning rates for each expert in a finite set. Those learning rates are computed with a second order refinement of losses (or a linearized version when the loss is convex in its second argument) for each expert, at each time round. The regret of his strategy with respect to best aggregation of $M$ finite experts is of the order of $\log M\sqrt{T}\log \log T$.  However, the context for this procedure is not the same as ours, as we resort to the Gibbs quasi-posterior which is defined on $\mathcal{C}$, a continuous set. In addition, we focus on a single temperature parameter $\lambda$ for the sake of computational complexity since the second order refinement requires the computation of the expectation of loss with respect to each expert in a finite set while, in our case, the "expert set" (\emph{i.e.,} $\mathcal{C}$) is continuous, leading to the tedious computation of second order refinement.

\subsection{Adaptive regret bounds}

The time horizon $T$ is usually unknown, prompting us to choose a time-dependent inverse temperature parameter $\lambda=\lambda_{t}$.
We thus propose a generalization of \autoref{proc:procedure 1}, described in \autoref{proc:alg1bis}.

\begin{algorithm}[h]
\caption{The adaptive quasi-Bayesian online clustering algorithm}
\label{proc:alg1bis}
\begin{algorithmic}[1]
\\\textbf{Input parameters}: $p>0,\pi\in\mathcal{P}(\mathcal{C})$, $(\lambda_{t})_{0:T}>0$ and $S_{0}\equiv 0$
\\\textbf{Initialization}: Draw $\hat{\bc}_{1} \sim \pi=\hat{\rho}_1$
\State \textbf{For $t\in\llbracket 1,T\rrbracket$}
\State\hspace{\algorithmicindent} Get the data $x_{t}$
\State \hspace{\algorithmicindent} Draw $\hat{\bc}_{t+1} \sim \hat{\rho}_{t+1}(\bc)$ where $\mathrm{d}\hat{\rho}_{t+1}(\bc)\propto\exp\big(-\lambda_{t}S_{t}(\bc)\big)\mathrm{d}\pi(\bc)$,
and
\begin{equation*}
S_{t}(\bc)=S_{t-1}(\bc)+\ell(\bc,x_{t})+\frac{\lambda_{t-1}}{2}\big(\ell(\bc,x_{t})-\ell(\hat{\bc}_{t},x_{t})\big)^2.
\end{equation*}
\State \textbf{End for}
\end{algorithmic}
\end{algorithm}

This adaptive algorithm is supported by the following more involved regret bound.
\begin{theo}\label{thm:theorem 2}
For any sequence $(x_{t})_{1:T}\in \R^{dT}$, any prior distribution $\pi$ on $\mathcal{C}$, if $(\lambda_{t})_{0:T}$ is a non-increasing sequence of positive numbers, then the procedure described in \autoref{proc:alg1bis} satisfies
\begin{multline*}
\expcum\leq \inf_{\rho \in \PP_\pi(\mathcal{C})}\left\{\mathbb{E}_{\bc\sim \rho}\left[\sum_{t=1}^{T}\ell(\bc,x_{t})\right]+\frac{\K(\rho,\pi)}{\lambda_{T}} \right. \\ \left.
+\mathbb{E}_{(\hat{\rho}_{1},\dots,\hat{\rho}_{T})}\mathbb{E}_{\bc\sim \rho}\left[\sum_{t=1}^{T}\frac{\lambda_{t-1}}{2}[\ell(\bc,x_{t})-\ell(\hat{\bc}_{t},x_{t})]^2\right]\right\}.
\end{multline*}
\end{theo}
If $\lambda$ is chosen in \autoref{thm:theorem 1} as $\lambda=\lambda_{T}$, the only difference between \autoref{thm:theorem 1} and \autoref{thm:theorem 2} lies on the last term of the regret bound. This term will be larger in the adaptive setting than in the simpler non-adaptive setting since $(\lambda_{t})_{0:T}$ is non-increasing. In other words, here is the price to pay for the adaptivity of our algorithm. However, a suitable choice of $\lambda_{t}$ allows, again, for a refined result.
\begin{coro}\label{cor:corollary 3}
For any deterministic sequence $(x_{t})_{1:T}\in \R^{dT}$, if $q$ and $\pi_{k}$ in \eqref{eq:prior distribution} are taken respectively as in \eqref{eq:prior on finite set} and \eqref{eq:uniform prior} with $\eta \geq 0$ and $R\geq \max_{t=1,\dots,T}|x_{t}|_{2}$, if $\lambda_{t}=(d+2)\sqrt{\log t}/\left(2\sqrt{t}R^{2}\right)$ for any $t \in \llbracket1,T\rrbracket$ and $\lambda_{0}=1$, then for $T\geq 5$ the procedure described in \autoref{proc:alg1bis} satisfies
\begin{multline*}
\sum_{t=1}^T\mathbb{E}_{(\hat{\rho}_{1},\hat{\rho}_{2},\dots,\hat{\rho}_{t})}\ell(\hat{\bc}_{t},x_{t})\leq \inf_{k\in \llbracket 1,p\rrbracket}\left\{\inf_{\bc\in\mathcal{C}(k,R)}\sum_{t=1}^{T}\ell(\bc,x_{t})+\frac{2(d+\eta)R^{2}}{d+2}k\sqrt{T\log T}\right\} \\+\left(  \frac{2R^{2}\log p}{d+2}+\frac{dR^{2}}{d+2}+\frac{81(d+2)R^{2}}{2}\right)\sqrt{T \log T}.
\end{multline*}
\end{coro}
Therefore, the price to pay for not knowing the time horizon $T$ (which is a much more realistic assumption for online learning) is a multiplicative factor $2$ in front of the term $\frac{81(d+2)R^{2}}{4}\sqrt{T \log T}$. This does not degrade the rate of convergence $\sqrt{T\log T}$.
%The adaptive algorithm described in \autoref{proc:alg1bis} is therefore supported by a regret bound with rate $\sqrt{T}\log T$.

\medskip

In the next corollary, we use the doubling trick (\citealp[][Section 2.3]{CBL2006}, also appearing in \citealp{CMS2007}) to show how can we overcome the difficulty when a priori bound $R$ on the $\ell_{2}$-norm of sequence $(x_{t})_{1:T}$ is unknown.

\medskip
Let us first denote by $R_{0} = 1$, and for $t\geq 1$
\begin{equation*}
    R_{t} = \max_{s=1,\dots,t}2^{\ceil*{\log_{2}\left(\left|x_{s}\right|_{2}\right)}},
\end{equation*}
where $\ceil*{x}$ represents the least integer greater than or equal to $x \in \R$. It is easy to see that $(R_{t})_{t\geq 1}$ is non-decreasing and satisfies for any $t\geq 1$ $$\max_{s=1,\dots, t}\left|x_{s}\right|_{2} \leq R_{t}\leq 2 \max_{s=1,\dots, t}\left|x_{s}\right|_{2}.$$ We call epoch $r$, $r=0,1,\dots,$ the sequence $(t_{r-1}+1, t_{r-1}+2, \dots, t_{r})$ of time steps where the last step $t_{r}$ is the time step $t=t_{r}$ when $R_{t} > R_{t_{r-1}}$ take places for the first time (we set conventionally $t_{-1} = 0$). Within each epoch $r\geq 0$, \emph{i.e.,} for $t \in \left[t_{r-1}+1, \dots, t_{r}\right]$, let
\begin{equation*}
    \lambda_{r,t} = \frac{(d+2)\sqrt{\log t}}{2\sqrt{t}R_{t_{r-1}}^{2}}.
\end{equation*}

Let \textbf{Alg-R} be a prediction algorithm that runs \autoref{proc:alg1bis} in each epoch $r$ with parameter $\lambda_{r,t}$, then we have the following result.

\begin{coro}\label{cor:corollary 4}
For any deterministic sequence $(x_{t})_{1:T}\in \R^{dT}$, if $q$ and $\pi_{k}$ in \eqref{eq:prior distribution} are taken respectively as in \eqref{eq:prior on finite set} and \eqref{eq:uniform prior} with $\eta \geq 0$, the regret of algorithm $\mathbf{Alg}$-$\mathbf{R}$ satisfies
\begin{multline*}
\sum_{t=1}^T\mathbb{E}_{(\hat{\rho}_{1},\hat{\rho}_{2},\dots,\hat{\rho}_{t})}\ell(\hat{\bc}_{t},x_{t})\leq \inf_{k\in \llbracket 1,p\rrbracket}\left\{\inf_{\bc\in\mathcal{C}(k,R)}\sum_{t=1}^{T}\ell(\bc,x_{t})+\frac{56(d+\eta)R^{2}}{3(d+2)}k\sqrt{T\log T}\right\} \\+\frac{28}{3}\left(\frac{2R^{2}\log p}{d+2}+\frac{dR^{2}}{d+2}+\frac{81(d+2)R^{2}}{2}\right)\sqrt{T \log T} + \frac{112}{3}R^{2},
\end{multline*}
where $R=\max_{t=1,\dots,T}|x_{t}|_{2}$.
\end{coro}
Note that the price to pay for making our algorithm adaptive to unknown bound $R$ is a multiplicative term $\frac{28}{3}$ and an additional $\frac{112}{3}R^{2}$ in the regret bound.

\subsection{Minimax regret}\label{sec:minimax}
This section is devoted to the study of the minimax optimality of our approach. The regret bound in \autoref{cor:corollary 3} has a rate $\sqrt{T \log T}$, which is not a surprising result. Indeed, many online learning problems give rise to similar bounds depending also on the properties of the loss function. However, in the online clustering setting, it is legitimate to wonder wether the upper bound is tight, and more generally if there exists other algorithms which provide smaller regrets. The sequel answers both questions in a minimax sense.
\medskip

Let us first denote by $\left|\bc \right|$ the number of cells for a partition $\bc \in \mathcal{C}$. We also introduce the following assumption.
\medskip

\textbf{Assumption $\mathcal{H}(s)$}: \textit{Let $R>0$ and $T\in \mathbb{N}^{\ast}$. For a given $s\in \llbracket 1, p\rrbracket$, we assume that the number of cells $\left|\bc_{T,R}^\star\right|$ for partition $\bc_{T,R}^\star$ defined by the following
\begin{equation*}
  \bc_{T,R}^\star=\underset{\bc\in \cup_{k=1}^{p}\mathcal{C}(k, R)}{\arg\min}\left\{\sum_{t=1}^T\ell(\bc,x_t)+|\bc|\sqrt{T\log T} \right\}.
\end{equation*}
equals to $s$, \emph{i.e.,}$\left|\bc_{T,R}^\star\right| =s$. 
% is defined in \autoref{sec:notation}.
}
\medskip

Note that several partitions may achieve the minimum. In that case, we adopt the convention that $\bc_{T,R}^{\star}$ is any such partition with the smallest number of cells. Assumption $\mathcal{H}(s)$ means that $(x_t)_{1:T}$ could be well summarized by $s$ cells since the infimum is reached for the partition $\bc^\star_{T,R}$. We introduce the set
\begin{equation*}
  \omega_{s,R}=\Big\{\left(x_t\right)\mbox{ such that }\mathcal{H}(s) \mbox{ holds}\Big\}\subseteq \R^{dT}.
\end{equation*}
For \autoref{proc:alg1bis}, we have from \autoref{cor:corollary 3} that
\begin{align*}
\sup_{(x_t)\in\omega_{s,R}}\left\{\expcum-\inf_{\bc\in\CC(s,R)}\sum_{t=1}^T\ell(\bc,x_t)\right\}\leq c_{1}\times s \sqrt{T \log T},
\end{align*} 
where $c_{1}$ is a constant depending on $R, d, p$ (recall that they are respectively the bound on the $\ell_{2}$-norm of sequence $(x_{t})_{1:T}$, the dimension of the data point and the maximum number of cells allowed for clustering). 
\medskip

Then for any $s\in\mathbb{N}^{\ast}$, $R>0$, our goal is to obtain a lower bound of the form
\begin{equation*}
    \inf_{(\hat{\rho}_{t})}\ \sup_{(x_t)\in\omega_{s,R}}\left\{\expcum-\inf_{\bc\in\CC(s,R)}\sum_{t=1 }^T\ell(\bc,x_t)\right\}\geq c_{2}\times s\sqrt{T \log T},
\end{equation*}
where $c_{2}$ is some constant satisfying $c_{2}\leq c_{1}$. 
\medskip

The first infimum is taken over all distributions $(\hat{\rho}_{t})_{1:T}$ whose support is  $\cup_{k=1}^{p}\prod_{j=1}^{k}\mathit{B}_{d}(2R)$, where $\mathit{B}_{d}(2R)$ is defined in \eqref{ball}. 
Next, we obtain
\begin{multline}\label{ieq:prob setting}
 \inf_{(\hat{\rho}_{t})}\ \sup_{(x_t)\in\omega_{s,R}}\left\{\expcum-\inf_{\bc\in\CC(s,R)}\sum_{t=1 }^T\ell(\bc,x_t)\right\} \\ \geq\inf_{(\hat{\rho}_{t})} \hspace{0.1cm}\E_{\mu^T}\left\{\rdexpcum 
 - \inf_{\bc \in \CC(s,R)}\sum_{t=1 }^T\ell(\bc,X_t)\right\}\mathbb{1}_{\left\{(X_{t}) \in \omega_{s,R})\right\}},
%\E_{\nu^N}\min_{j=1,\ldots, N}\sum_{t=1 }^T\ell(\bc_j,X_t)\right\},   
\end{multline}
where $X_t$, $t=1, \ldots, T$  are i.i.d with distribution $\mu$ defined on $\R^{d}$ and $\mu^{T}$ stands for the joint distribution of $(X_{1},\dots,X_{T})$. Unfortunately, in \eqref{ieq:prob setting}, since the infimum is taken over any distribution $(\hat{\rho}_{t})$, there is no restriction on the number of cells of each partition $\hat{\bc}_{t}$ . Then, the left hand side of \eqref{ieq:prob setting} could be arbitrarily small or even negative and the lower bound does not match the upper bound of \autoref{cor:corollary 3}. To handle this, we need to introduce a penalized term which  accounts for the number of cells of each partition to the loss function $\ell$. The upcoming theorem provides minimax results for an augmented value $\mathcal{V}_{T}(s)$ defined as
\begin{equation}\label{eq:minimax notation}
\mathcal{V}_T(s)=\inf_{(\hat{\rho}_{t})}\ \sup_{(x_t)\in\omega_{s,R}}\left\{\sum_{t=1}^T\mathbb{E}_{(\hat{\rho}_{1},\dots,\hat{\rho}_{t})}\left(\ell(\hat{\bc}_t,x_t)+\frac{\sqrt{\log T}}{\sqrt{T}}\left|\hat\bc_t\right|\right)-\inf_{\bc\in \CC(s,R)}\sum_{t=1 }^T\ell(\bc,x_t)\right\}.
\end{equation}
In \eqref{eq:minimax notation}, we add a term which penalizes the number of cells of each partition. To capture the asymptotic behavior of $\mathcal{V}_T(s)$, we derive an upper bound for the penalized loss in \eqref{eq:minimax notation}. This is done in the following theorem which combines an upper and lower bound for the regret, hence proving that it is minimax optimal.
\begin{theo}
\label{thm:theorem 3}
Let $s\in\mathbb{N}^{\ast}$, $R>0$ such that
\begin{align}
\label{conds}
2\leq s\leq \left\lfloor\left(\frac{RT^{\frac{1}{4}}}{6\log T^{\frac{1}{4}}}\right)^{\frac{d}{d+1}}\right\rfloor,
%\left\lfloor\left(\frac{RT^{\frac{1}{4}}}{\sqrt{\log T}}\right)^{\frac{d}{2d+1}}\right\rfloor + 1.
\end{align}
where $\lfloor x \rfloor$ represents the largest integer that is smaller than $x$.
If $T$ satisfies $T^{\frac{d+2}{2}} \geq 8R^{2d}\sqrt{\log T}$, then
\begin{align}
\label{lbonline}
s\sqrt{T\log T} \left(1-\frac{2}{T}\left[1+\frac{s-1}{2s^2}\right]\right)\leq \mathcal{V}_T(s)\leq \mathrm{const.}\times s\sqrt{T\log T}.
\end{align}
\end{theo}
The lower bound on $\mathcal{V}_T(s)/T$ is asymptotically of order $\sqrt{\log T}/\sqrt{T}$. Note that \cite{BLL1998} obtained the less satisfying rate $1/\sqrt{T}$, however holding with no restriction on the number of cells retained in the partition whereas our claim has to comply with \eqref{conds}. This is the price to pay for our additional $\sqrt{\log T}$ factor. Note however that this price is mild as $s$ is no longer upper bounded whenever $T$ or $R$ grow to $+\infty$, casting our procedure onto the online setting where the time horizon is not assumed finite and the number of clusters is evolving along time.
\medskip

As a conclusion to the theoretical part of the manuscript, let us summarize our results. Regret bounds for \autoref{proc:procedure 1} are produced for our specific choice of prior $\pi$ (\autoref{cor:corollary 1})
and with an involved choice of $\lambda$ (\autoref{cor:corollary 2}). For the adaptive version \autoref{proc:alg1bis}, the pivotal result is \autoref{thm:theorem 2}, which is instantiated for our prior in \autoref{cor:corollary 3}. Finally, the lower bound is stated in \autoref{thm:theorem 3}, proving that our regret bounds are minimax whenever the number of cells retained in the partition satisfies \eqref{conds}. We now move to the implementation of our approach.

\section{The \name\ algorithm}\label{sec:algo}

Since direct sampling from the Gibbs quasi-posterior is usually not possible, we focus on a stochastic approximation in this section, called \name\ \citep[available in the companion eponym R package from][]{R-PACBO}. Both implementation and convergence (towards the Gibbs quasi-posterior) of this scheme are discussed. This section also includes a short numerical experiment on synthetic data to illustrate the potential of \name\ compared to other popular clustering methods.

\subsection{Structure and links with RJMCMC}

In \autoref{proc:procedure 1} and \autoref{proc:alg1bis}, it is required to sample at each $t$ from the Gibbs quasi-posterior $\hat{\rho}_t$. Since $\hat{\rho}_t$ is defined on the massive and complex-structured space $\mathcal{C}$ (let us recall that $\mathcal{C}$ is a union of heterogeneous spaces), direct sampling from $\hat{\rho}_t$ is not an option and is much rather an algorithmic challenge. Our approach consists in approximating $\hat{\rho}_t$ through MCMC under the constraint of favouring local moves of the Markov chain. To do it, we will use resort to Reversible Jump MCMC \citep{Gre1995}, adapted with ideas from the Subspace Carlin and Chib algorithm proposed by \cite{DFN2002} and \cite{PD2012}.
 %and our implementation is close in spirit to the one of \cite{GA2013} and \cite{GR2015}.
 Since sampling from $\hat{\rho}_t$ is similar for any $t=1,\dots,T$, the time index $t$ is now omitted for the sake of brevity.
\medskip

\begin{figure}[h]
\begin{equation*}
\begin{tikzpicture}
\node at (11,6.5) {$\bc^{(n)}$};
\node at (7,6.5) {$k^{(n)}$};
\draw [very thick](10.5,6) rectangle (11.5,7);
\draw [very thick](6.5,6) rectangle (7.5,7);
\draw [-triangle 60](7,6)--(6.1,5);
\node [left] at (6.2,5.7) {$q(k^{(n)},\cdot)$};
\node at (5.8,4.5) {$k^{\prime}$};
\draw [red,thick,dashed](5.8,4.5) circle (0.5cm);
\draw [-triangle 60](6.3,4.5)--(7.8,4.5);
\node [above] at (7,4.5) {$\rho_{k^{\prime}}(\cdot)$};
\node at (8.3,4.5) {$v_{1}$};
\draw [thick] (8.3,4.5) circle(0.5cm);

\draw [-triangle 60](8.7,4.1)--(9.2,3.6);
\draw [-triangle 60](9.296,3.47)--(10.7,2);
\draw [-triangle 60](11,6)--(9.4,3.6);
\draw [-triangle 60](9.296,3.47)--(8.3,2);

\node [right] at (9.5,3.5) {$g\left(v_{1},\bc^{(n)}\right)$};
\filldraw[color=black!60, fill=black!5, very thick](9.3,3.5) circle (0.1cm);
\node at (8.1,1.5) {$v_{2}$};
\node at (11,1.5) {$\bc^{\prime}$};
\draw [red, thick, dashed] (11,1.5) circle(0.5cm);
\draw [thick] (8.1,1.5) circle(0.5cm);

\draw [-triangle 60](6.5,6.5)--(4,6.5)--(4,0)--(6.2,0);

\node at (7.3,0) {$k^{(n+1)}=k^{\prime}$};
\node [above,minimum width=0.1cm, minimum height=0.1cm] at (5,0) {\small w.p. $\alpha$};
\draw [very thick](6.2,-0.5) rectangle (8.6,0.5);

\draw [-triangle 60](4,0)--(4,-1.5)--(6.2,-1.5);
\node [above,minimum width=0.1cm, minimum height=0.1cm] at (5,-1.5) {\small w.p. 1-$\alpha$};
\node at (7.4,-1.5) {$k^{(n+1)}=k^{(n)}$};
\draw [very thick](6.2,-2) rectangle (8.6,-1);

\draw [-triangle 60](11.5,6.5)--(14,6.5)--(14,0)--(12.2,0);
\node at (11,0) {$\bc^{(n+1)}=\bc^{\prime}$};
\draw [very thick](9.8,-0.5) rectangle (12.2,0.5);
\node [above, minimum width=0.1cm, minimum height=0.1cm] at (13,0) {\small w.p. $\alpha$};

\draw [- triangle 60](14,0)--(14,-1.5)--(12.2,-1.5);
\node [above,minimum width=0.1cm, minimum height=0.1cm] at (13,-1.5) {\small w.p. 1-$\alpha$};
\draw [very thick](9.8,-2) rectangle (12.2,-1);
\node at (11,-1.5) {$\bc^{(n+1)}=\bc^{(n)}$};

\end{tikzpicture}
\end{equation*}
\caption{General structure of \name.}\label{Fi:structure}
\end{figure}
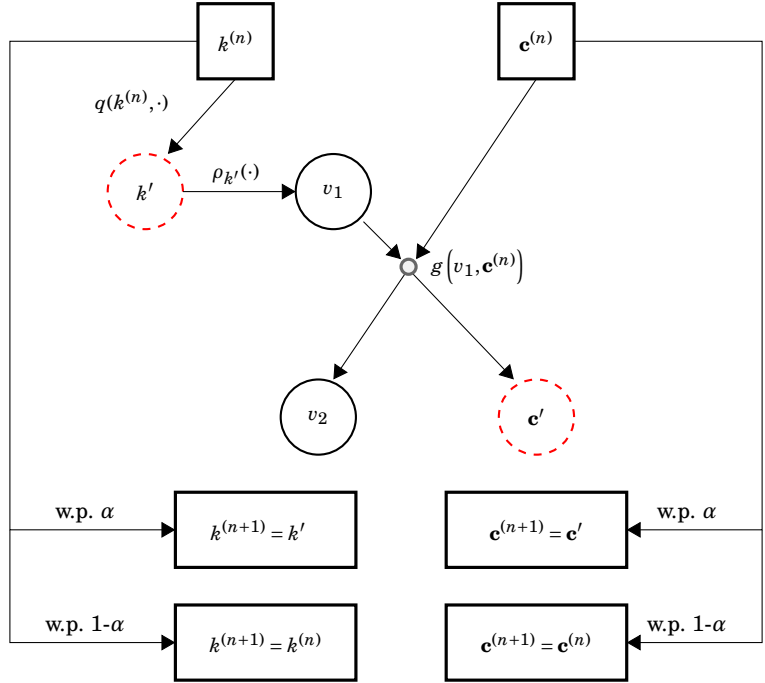

Let $(k^{(n)},\bc^{(n)})_{0\leq n\leq N}$, $N\geq 1$ be the states of the Markov Chain of interest of length $N$, where $k^{(n)}\in \llbracket 1,p \rrbracket$ and $\bc^{(n)}\in \Rkn$. At each RJMCMC iteration, only local moves are possible from the current state $(k^{(n)},\bc^{(n)})$ to a proposal state $(k^{\prime},\bc^{\prime})$, in the sense that the proposal state should only differ from the current state by at most one covariate. Hence, $\bc^{(n)}\in \R^{dk^{(n)}}$ and $\bc^{\prime} \in \R^{dk^{\prime}}$ may be in different spaces ($k^{\prime}\neq k^{(n)}$). Two auxiliary vectors $v_{1}\in \mathbb{R}^{d_{1}}$ and $v_{2}\in \mathbb{R}^{d_{2}}$ with $d_{1},d_{2}\geq 1$ are needed to compensate for this dimensional difference, \ie, satisfying the dimension matching condition introduced by \cite{Gre1995}
\begin{equation*}
dk^{(n)}+d_{1}=dk^{\prime}+d_{2},
\end{equation*}
such that the pairs $(v_{1},\bc^{(n)})$ and $(v_{2},\bc^{\prime})$ are of analogous dimension. This condition is a preliminary to the detailed balance condition that ensures that the Gibbs quasi-posterior $\hat{\rho}_t$ is the invariant distribution of the Markov chain. The structure of \name\ is presented in \autoref{Fi:structure}.

\medskip
Let $\rho_{k^{\prime}}(\cdot,\ttc_{k^{\prime}},\tau_{k^{\prime}})$ denote the multivariate Student distribution on $\mathbb{R}^{dk^{\prime}}$
\begin{equation}\label{eq:proposal distribution}
\rho_{k^{\prime}}(\bc,\ttc_{k^{\prime}},\tau_{k^{\prime}})=\prod_{j=1}^{k^{\prime}}\left\{C_{\tau_{k^\prime}}^{-1}\left(1+\frac{|c_{j}-\ttc_{k^{\prime},j}|^2_{2}}{6\tau_{k^{\prime}}^2}\right)^{-\frac{3+d}{2}}\right\}\mathrm{d}\bc,
\end{equation}
where $C_{\tau_{k^\prime}}^{-1}$ denotes a normalizing constant. Let us now detail the proposal mechanism. First, a local move from $k^{(n)}$ to $k^{\prime}$ is proposed by choosing $k^{\prime} \in \llbracket k^{(n)}-1,k^{(n)}+1\rrbracket$ with probability $q(k^{(n)},\cdot)$. Next, choosing $d_{1}=dk^\prime$, $d_{2}=dk^{(n)}$, we sample $v_{1}$ from $\rho_{k^{\prime}}$ in \eqref{eq:proposal distribution}. Finally, the pair $(v_{2},\bc^{\prime})$ is obtained by
\begin{equation*}
(v_{2},\bc^{\prime})=g\left(v_{1},\bc^{(n)}\right),
\end{equation*}
where $g: (x,y)\in \mathbb{R}^{dk^{\prime}} \times \mathbb{R}^{dk^{(n)}}\mapsto (y,x)\in \mathbb{R}^{dk^{(n)}} \times \mathbb{R}^{dk^{\prime}}$ is a one-to-one, first order derivative mapping. The resulting RJMCMC acceptance probability is
\begin{align*}%\label{eq:acceptance probability}
\alpha \left[\left(k^{(n)},\bc^{(n)}\right),\left(k^{\prime},\bc^{\prime}\right)\right] 
&=\min \left\{1,\frac{\hat{\rho}_{t}(\bc^{\prime})q(k^{\prime},k^{(n)})\rho_{k^{(n)}}(v_{2})}{\hat{\rho}_{t}(\bc^{(n)})q(k^{(n)},k^{\prime})\rho_{k^{\prime}}(v_{1})}
\left|\frac{\partial g\left(v_{1},\bc^{(n)}\right)}{\partial v_{1}\partial \bc^{(n)}}\right|\right\}, \\
&=\min\left\{1,\frac{\hat{\rho}_{t}(\bc^{\prime})q(k^{\prime},k^{(n)})\rho_{k^{(n)}}(\bc^{(n)},\ttc_{k^{(n)}},\tau_{k^{(n)}})}{\hat{\rho}_{t}(\bc^{(n)})q(k^{(n)},k^{\prime})\rho_{k^{\prime}}(\bc^{\prime},\ttc_{k^{\prime}},\tau_{k^{\prime}})}\right\},
\end{align*}
since the determinant of the Jacobian matrix of $g$ is $1$. The resulting \name\ algorithm is described in \autoref{proc:procedure 2}.

\begin{algorithm}[h]
\caption{\name}
\begin{algorithmic}[1]
\State \text{\textbf{Initialization}:} $(\lambda_t)_{1:T}$\\
\text{\textbf{For} $t \in \llbracket 1,T\rrbracket$}
\State \text{\textbf{Initialization}:} $\left(k^{(0)},\bc^{(0)}\right) \in \llbracket 1,p \rrbracket\times \R^{dk^{(0)}}$. Typically $k^{(0)}$ is set to $k^{(N)}$ from iteration $t-1$ ($k^{(0)}=1$ at iteration $t=1$).\\
\textbf{For} $n\in\llbracket 1,N-1\rrbracket$
\State \hspace{\algorithmicindent} Sample $k^{\prime}\in  \left\llbracket \max(1,k^{(n)}-1),\min(p,k^{(n)}+1)\right\rrbracket$ from $q(k^{(n)},\cdot)=\frac{1}{3}$.
\State \hspace{\algorithmicindent} Let $\ttc^{\prime} \gets  \text{standard $k^\prime$-means output}$ trained on $(x_{s})_{1:(t-1)}$.
\State \hspace{\algorithmicindent} Let $\tau^{\prime}= 1/\sqrt{pt}.$
\State \hspace{\algorithmicindent} Sample $v_{1}$ $\sim$ $\rho_{k^{\prime}}(\cdot,\ttc_{k^{\prime}},\tau_{k^{\prime}})$.
\State \hspace{\algorithmicindent} Let $(v_{2},\bc^{\prime})=g(v_{1},\bc^{(n)}).$
\State \hspace{\algorithmicindent} Accept the move $(k^{(n)},\bc^{(n)})=(k^{\prime},\bc^{\prime})$ with probability  
\begin{align*}
 \alpha\left[(k^{(n)},\bc^{(n)}),(k^{\prime}, \bc^{\prime}))\right]
=&\min\left\{1,\frac{\hat{\rho}_{t}(\bc^{\prime})q(k^{\prime},k^{(n)})\rho_{k^{(n)}}(v_{2},\ttc_{k^{(n)}},\tau_{k^{(n)}})}{\hat{\rho}_{t}(\bc^{(n)})q(k^{(n)},k^{\prime})\rho_{k^{\prime}}(v_{1},\ttc_{k^{\prime}},\tau_{k^{\prime}})}
\left|\frac{\partial g(v_{1},\bc^{(n)})}{\partial v_{1}\partial \bc^{(n)}}\right|\right\}\\
=&\min\left\{1,\frac{\hat{\rho}_{t}(\bc^{\prime})q(k^{\prime},k^{(n)})\rho_{k^{(n)}}(\bc^{(n)},\ttc_{k^{(n)}},\tau_{k^{(n)}})}{\hat{\rho}_{t}(\bc^{(n)})q(k^{(n)},k^{\prime})\rho_{k^{\prime}}(\bc^{\prime},\ttc_{k^{\prime}},\tau_{k^{\prime}})}\right\}
\end{align*}
\State \hspace{\algorithmicindent} Else $(k^{(n+1)},\bc^{(n+1)})=(k^{(n)},\bc^{(n)}).$\\
\textbf{End for}
\State Let $\hat{\bc}_{t}=\bc^{(N)}.$\\
\textbf{End for}
\end{algorithmic}
\label{proc:procedure 2}
\end{algorithm}

\subsection{Convergence of \name\ towards the Gibbs quasi-posterior}

We prove that \autoref{proc:procedure 2} builds a Markov chain whose invariant distribution is precisely the Gibbs quasi-posterior as $N$ goes to $+\infty$. To do so, we need to prove that the chain is $\hat{\rho}_{t}$-irreducible, aperiodic and Harris recurrent, see \citet[][Theorem 6.51]{RG2004} and \citet[][Theorem 20]{RR2006}.
\medskip

Recall  that at each RJMCMC iteration in \autoref{proc:procedure 2}, the chain is said to propose a "between model move" if $k^{\prime}\neq k^{(n)}$ and a "within model move" if $k^{\prime}=k^{(n)}$ and $\bc^{\prime}\neq \bc^{(n)}$. The following result gives a sufficient condition for the chain to be Harris recurrent. 
\begin{lem}\label{lemma:convergence of rjmcmc}
Let $D$ be the event that no "within-model move" is ever accepted and $\mathcal{E}$ be the support of $\hat{\rho}_{t}$. Then the chain generated by \autoref{proc:procedure 2} satisfies
\begin{equation*}
\mathbb{P}\left[D|\left(k^{(0)},\bc^{(0)}\right)=(k,\bc)\right]=0,
\end{equation*}
for any $k \in \llbracket1,p\rrbracket $ and $\bc \in  \mathbb{R}^{dk}\cap \mathcal{E}$.
\end{lem}
\autoref{lemma:convergence of rjmcmc} states that the chain must eventually accept a "within-model move". It remains true for other choices of $q(k^{(n)},\cdot)$ in \autoref{proc:procedure 2}, provided that the stationarity of $\hat{\rho}_{t}$ is preserved.
\begin{theo}\label{thm:theorem 4}
Let $\mathcal{E}$ denote the support of $\hat{\rho}_{t}$. Then for any $\bc^{(0)} \in \mathcal{E}$, the chain $\left(\bc^{(n)}\right)_{1:N}$ generated by \autoref{proc:procedure 2} is $\hat{\rho}_{t}$-irreducible, aperiodic and Harris recurrent.
\end{theo}
\autoref{thm:theorem 4} legitimates our approximation \name\ to perform online clustering, since it asymptotically mimics the behavior of the computationally unavailable $\hat{\rho}_{t}$. To the best of our knowledge, this kind of guarantee is original in the PAC-Bayesian literature. 
\medskip

Finally, let us stress that obtaining an explicit rate of convergence is beyond the scope of the present work. However, in most cases the chain converges rather quickly in practice, as illustrated by \autoref{fig:convergence of rjmcmc in practice}. At time $t$, we advocate for setting $k^{(0)}$ as $k^{(N)}$ from round $t-1$, as a warm start.

\subsection{Numerical study}\label{sec:simus}
This section is devoted to the illustration of the potential of our quasi-Bayesian approach on synthetic data. Let us stress that all experiments are reproducible, thanks to the \name\ R package \citep{R-PACBO}. We do not claim to be exhaustive here but rather show the (good) behavior of our implementation on a toy example.

\subsubsection{Calibration of parameters and mixing properties}

\begin{figure}[h]
    \begin{subfigure}{.5\linewidth}
    \centering
    \includegraphics[width=\textwidth, height=3cm]{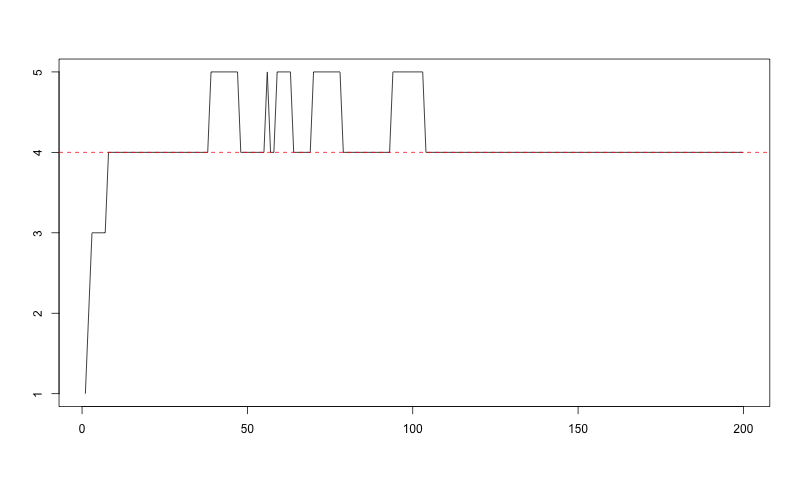}\hfill
    \caption{Number of clusters.}
  \end{subfigure}
  \begin{subfigure}{.5\linewidth}
    \centering
    \includegraphics[width=\textwidth, height=3cm]{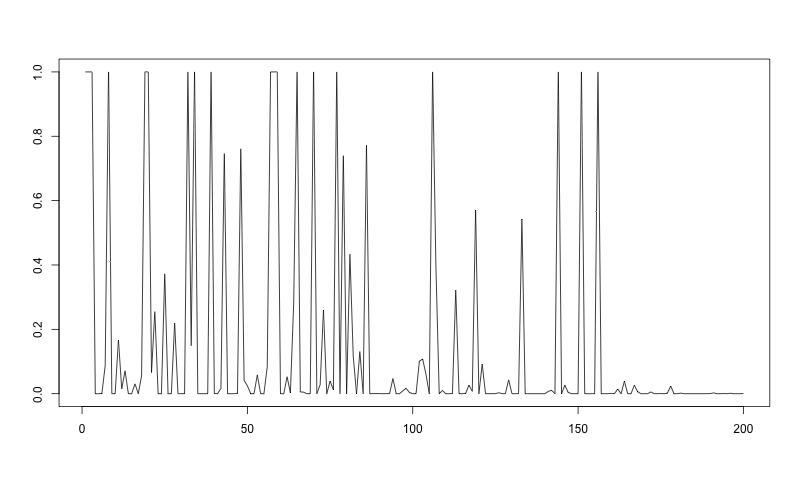}\hfill
    \caption{Acceptance probability.}
  \end{subfigure}  
  \caption{Typical RJMCMC output in \name. (a) $k^{(n)}_{1:N}$, number of clusters along the 200 iterations. The true number of clusters (set to $4$ in this example) is indicated by a dashed red line (b) acceptance probability $\alpha$ along the 200 iterations, exhibiting its mixing behavior.}
    \label{fig:convergence of rjmcmc in practice}
\end{figure}

We set $R$ to be the maximum $\ell_{2}$-norm of the observations. Note that a too small value will yield acceptance ratios to be close to zero and will degrade the mixing of the chain. As advised by the theory, we advise to set $\lambda_{t}=0.6\times(d+2)\sqrt{\log t}/(2\sqrt{t})$. Recall that large values will enforce the quasi-posterior to account more for past data, whereas small values make the quasi-posterior alike the prior. We illustrate in \autoref{fig:convergence of rjmcmc in practice} the mixing behavior of \name. The convergence occurs quickly, and the default length of the RJMCMC runs is set to 500 in the \name\ package: this was a ceiling value in all our simulations.

\subsubsection{Batch clustering setting}

A large variety of methods have been proposed in the literature for selecting the number $k$ of clusters in batch clustering \citep[see][for a survey]{MC1985,Gor1999}. These methods may be of local or global nature. For local methods, at each step, each cluster is either merged with another one, split in two or remains. Global methods evaluate the empirical distortion of any clustering as a function of the number $k$ of cells over the whole dataset, and select the minimizer of this distortion. The rule of \cite{Har1975} is a well-known representative of local methods. Popular global methods include the works of \cite{CH1974}, \cite{KL1985} and \cite{KR1990}, where functions based on the empirical distortion or on the average of within-cluster dispersion of each point are constructed and the optimal number of clusters is the maximizer of these functions. In addition, the Gap Statistic \citep{TWH2001} compares the change in within-cluster dispersion with the one expected under an appropriate reference null distribution. More recently, CAPUSHE (CAlibrating Penalty Using Slope Heuristics) introduced by \cite{Fis2011} and \cite{BMM2012} addresses the problem from the penalized model selection perspective, in the form of two methods: DDSE (Data-Driven Slope Estimation) and Djump (Dimension jump). R packages implementing those methods are used with their default parameters in our simulations.
\medskip

In this section, we compare \name\ to the aforecited methods in a batch setting with $n=200$ observations simulated from the following 4 models.

\begin{model}[1 group in dimension 5]\label{mod:model 1}
Observations are sampled from a uniform distribution on the unit hypercube in $\mathbb{R}^{5}$.
\end{model}

\begin{model}[4 Gaussian groups in dimension 2]\label{mod:model 2}
Observations are sampled from 4 bivariate Gaussian distributions with identity covariance matrix, whose mean vectors are respectively $(0,0), (-2,-1), (0,4), (3,1)$. Each observation is uniformly drawn from one of the four groups.
\end{model}

%\begin{model}[4 strongly separated Gaussian groups in dimension 2]\label{mod:model 3} 
%Same as the previous model but with more separated mean vectors: $(0,0),(-4,-1),(0,7),(5,2)$. 
%\end{model}

\begin{model}[7 Gaussian groups in dimension 50]\label{mod:model 4}
Observations are sampled from 7 multivariate Gaussian distributions in $\mathbb{R}^{50}$ with identity covariance matrix, whose mean vectors are chosen randomly according to an uniform distribution on $[-10,10]^{50}$. Each observation is uniformly drawn from one of the seven groups.
\end{model}

\begin{model}[3 lognormal groups in dimension 3]\label{mod:model 5}
Observations are sampled from 3 multivariate lognormal distributions in $\mathbb{R}^{3}$ with identity covariance matrix, whose mean vectors are respectively $(1,1,1), (6,5,7), (10,9,11)$. Each observation is uniformly drawn from one of the three groups.
\end{model}

\begin{figure}
    \begin{subfigure}{\linewidth}
    \centering
    \includegraphics[width=14cm,height=6cm]{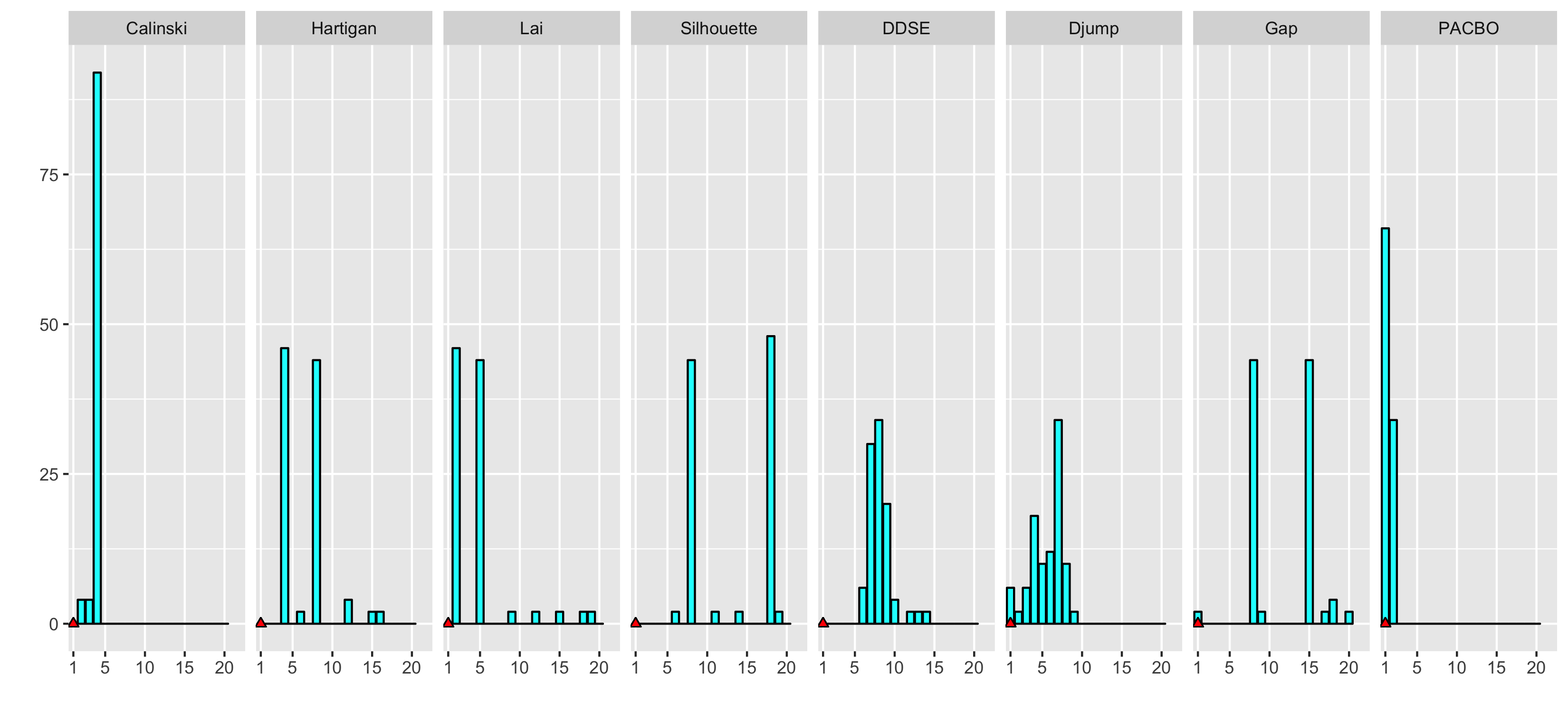}
    \caption{\autoref{mod:model 1}.}
    \label{Fig1a}
  \end{subfigure}
  \medskip
  \begin{subfigure}{\linewidth}
    \centering
    \includegraphics[width=14cm,height=6cm]{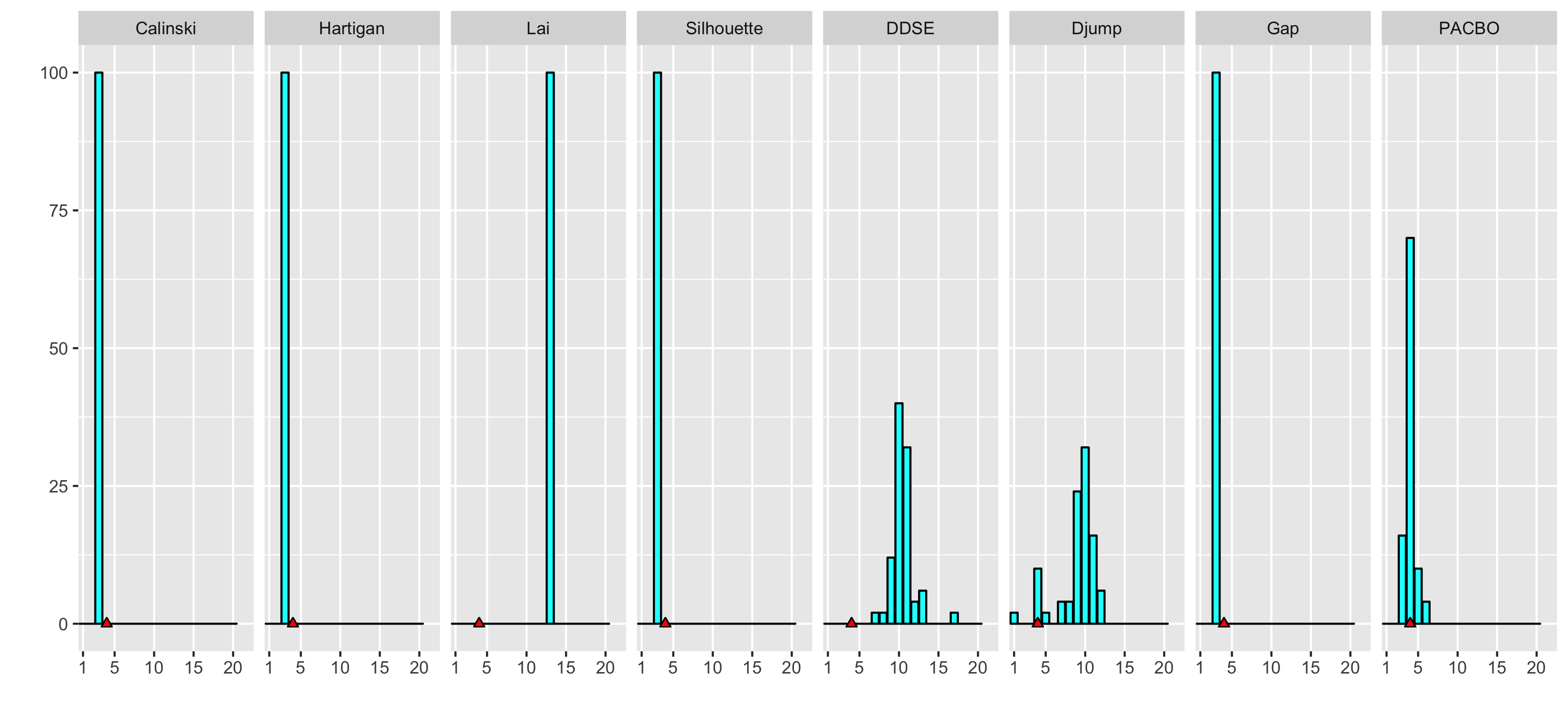}
    \caption{\autoref{mod:model 2}.}
    \label{Fig1b}
  \end{subfigure}
    \caption{Histograms of the estimated number of cells on 50 realizations. The red mark indicates the true number of cells.}
    \label{fig:Figure 1}
\end{figure}

\begin{figure}
      \begin{subfigure}{\linewidth}
    \centering
    \includegraphics[width=14cm,height=6cm]{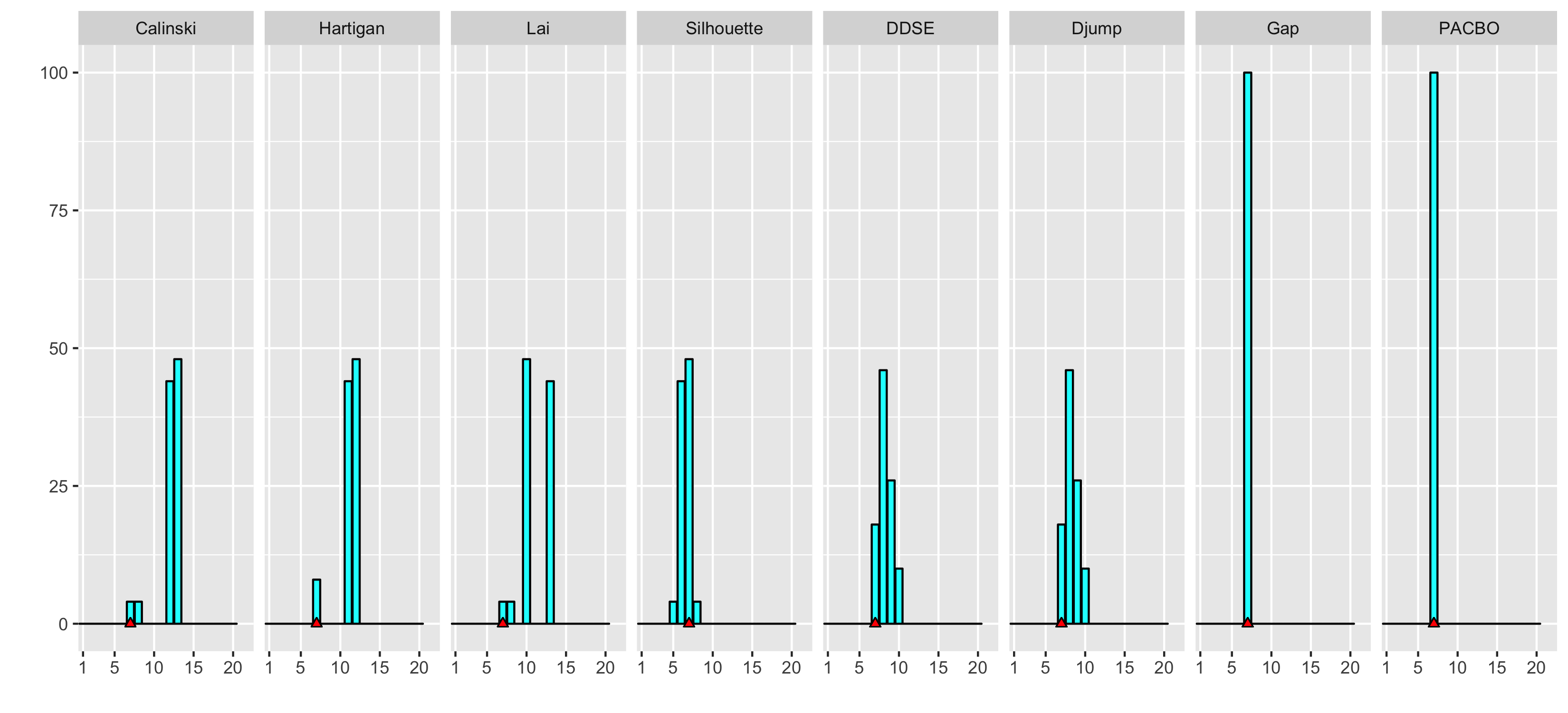}
    \caption{\autoref{mod:model 4}.}
    \label{Fig2b}
  \end{subfigure}
  \begin{subfigure}{\linewidth}
    \centering
    \includegraphics[width=14cm,height=6cm]{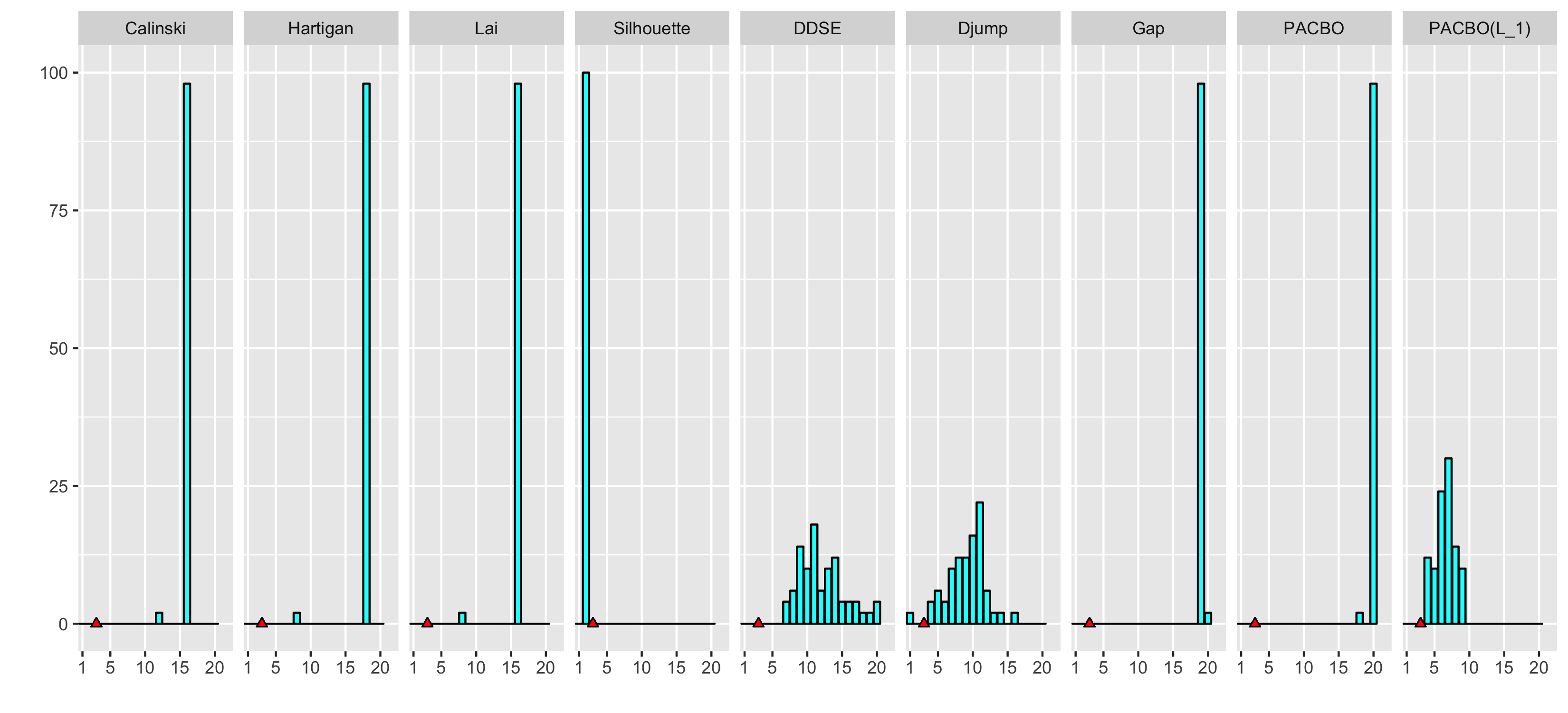}
    \caption{\autoref{mod:model 5}.}
    \label{Fig2c}
  \end{subfigure}
  \caption{Histograms of the estimated number of cells on 50 realizations. The red mark indicates the true number of cells.}
    \label{fig:Figure 2}
\end{figure}

\autoref{fig:Figure 1} and \autoref{fig:Figure 2} present the percentage of the estimated number of cells $k$ on 50 realizations of the 4 aforementioned models, for 8 methods including \name. In each graph, the red dot indicates the real number of groups. The methods used for selecting $k$ are presented on the top of each panel, where DDSE (Data-Driven Slope Estimation) and Djump (Dimension jump) are the two methods introduced in CAPUSHE \citep{BMM2012}. The maximum number of cells is set to 20.
\medskip

For \autoref{mod:model 1} \name\ outperforms all competitors, since it selects the correct number of cells in almost 70\% of our simulations, when all other methods barely find it
(\autoref{Fig1a}).

For \autoref{mod:model 2} Calinski, Hartigan, Silhouette and Gap underestimate the number of cells by identifying 3 groups. Djump finds the true value $k=4$ less than 10\%. \name\ identifies 4 groups in 60\% of our runs (\autoref{Fig1b}).

%For \autoref{mod:model 3}, \name\ does not perform as well as others ( \autoref{Fig1c}), illustrating the need for a fine tuning of $\lambda$. Indeed, dividing $\lambda$ by 3 yields far better results in (\autoref{Fig2a}).

For \autoref{mod:model 4} \name\ is one of the two best methods, together with Gap (\autoref{Fig2b}).

For \autoref{mod:model 5} where 3 groups of observations are generated from a heavy-tailed distribution, we consider a variant of \name\ with the $\ell_{1}$-norm in $\R^{d}$, \emph{i.e.}, we replace the loss in \eqref{eq:dynamic number of centers} by  $\ell(\hat{\bc}_{t},x_{t})=\min_{1\leq k\leq K_{t}}|\hat{c}_{t,k}-x_{t}|_{1}$. \autoref{Fig2c} shows that most methods perform poorly, to the notable exception of this \name ($\ell_1$).

\subsubsection{Online clustering setting}
In the last part, we have compared, in the batch setting, our method with 7 other methods on different datasets. However let us stress here that none of the aforementioned methods is specifically designed for \emph{online} clustering. Indeed, to the best of our knowledge \name\ is the sole procedure that explicitly takes advantage of the \emph{sequential nature} of data. For that reason, we present below the behavior and a comparison of running times between \name\ and the aforementioned methods, on the following synthetic online clustering toy example.
\begin{model}[10 mixed groups in dimension 2]\label{mod:model online}
Observations $(x_{t})_{t=1,\dots,T=200}$ are simulated in the following way: define firstly for each $t\in \llbracket1,T\rrbracket$ a pair $(c_{1,t},c_{2,t})\in \R^{2}$,  where $c_{1,t}= -\frac{5}{2}\pi + \frac{5\pi}{9}\left(\lfloor\frac{t-1}{20}\rfloor-1\right)$ and $c_{2,t}=5\sin(c_{1,t})$. Then for $t\in \left\llbracket1,100\right\rrbracket$, $x_{t}$ is sampled from a uniform distribution on the unit cube in $\mathbb{R}^{2}$, centered at $(c_{x,t},c_{y,t})$. For $t\in \llbracket101,200\rrbracket$, $x_{t}$ is generated by a bivariate Gaussian distribution, centered at $(c_{x,t},c_{y,t})$ with identity covariance matrix.
\end{model}\label{}
\medskip

\begin{figure}[h!]
  \begin{subfigure}{\linewidth}
    \centering
    \includegraphics[width=\textwidth, height=6cm]{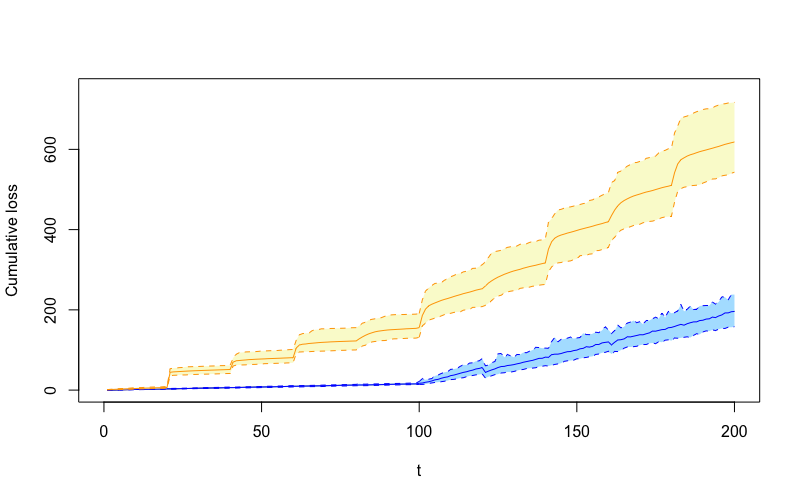}\hfill
    \caption{ECL (yellow line) and OCL (blue line) as function of $t$, with 95\% confidence intervals (dashed line).}
    \label{Fig4b}
  \end{subfigure}
  \begin{subfigure}{\linewidth}
    \centering
    \includegraphics[width=\textwidth, height=6cm]{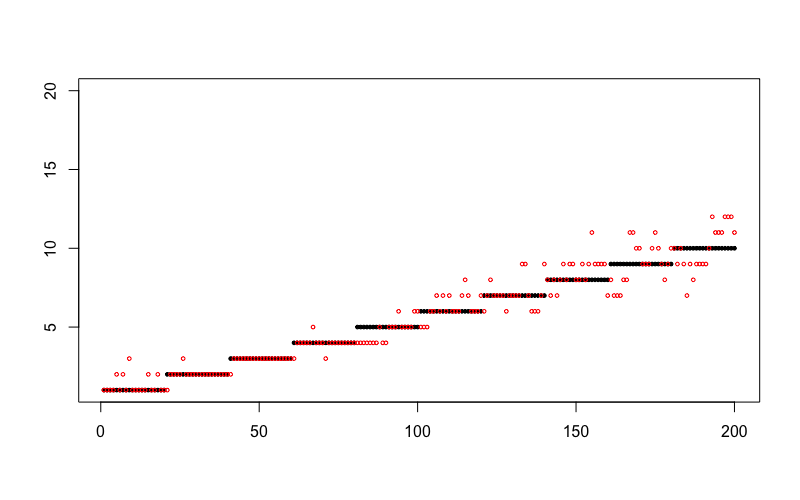}
    \caption{Estimated number of cells (red dots) by \name\ as a function of $t$. Black lines represent the true number of cells.}
    \label{Fig4c}
  \end{subfigure}
    \caption{Performance of \name.}
    \label{Fig:model_6_ECL_vs_OCL}
\end{figure}

In this online setting, the true number $k^\star_t$ of groups will augment of 1 unit every 20 time steps to eventually reach 10 (and the maximal number of clusters is set to $20$ for all methods). \autoref{Fig4b} shows  ECL for \name\ and OCL along with 95\% confidence intervals computed on 100 realizations with $T=200$ observations, with $\lambda_{t} = 0.6\times(d+2)/2\sqrt{t}$ and $R = 15$ (so that all observations are in the $\ell_2$-ball $B_{2}(R)$. Jumps in the ECL occur when new clusters of data are observed. Since \name\ outputs a partition based only on the past observations, the instantaneous loss is larger whenever a new cluster appears. However \name\ quickly identifies the new cluster. This is also supported by \autoref{Fig4c} which represents the true and estimated numbers of clusters.
\medskip

In addition we also count the number of correct estimations of the true number $k^\star_t$ of clusters. \autoref{table:estimation of number of clusters} contains its mean (and standard deviation, on $100$ repetitions) for \name\ and its seven competitors. \name\ has the largest mean by a significant margin and identifies the correct number of clusters of about 120 observations out of 200.

\begin{table}[h]
\begin{adjustbox}{max width=\textwidth}
\begin{tabular}{ cccccccccc}
%\hline
 Calinski & Hartigan & Lai & Silhouette & DDSE& Djump & Gap & \name  \\ \hline
  34.92 (8.24)  &  63.72 (4.81) & 52.23 (4.64) & 72.44 (4.39) & 22.73 (4.17) & 38.38 (6.21) & 56.73 (14.38) & \textbf{119.95 (7.08)}  \\ %\hline
%\hline
\end{tabular}
\end{adjustbox}
\caption{Mean and standard deviation of correct estimations of the true number of clusters.}
\label{table:estimation of number of clusters}
\end{table}
Next, we compare the running times of \name\ and its competitors, in the online setting. At each time $t=1,\dots,200$, we measure the running time of each method. \autoref{table:online execution time} presents the mean (and standard deviation) on $100$ repetitions of the total running times. The superiority of \name\ is a straightforward consequence of the fact that it adapts to the \emph{sequential nature} of data, whereas all other methods conduct a batch clustering at each time step.
\begin{table}[h]
\begin{adjustbox}{max width=\textwidth}
\begin{tabular}{ cccccccccc}
%\hline
 Calinski & Hartigan & Lai & Silhouette & DDSE & Djump & Gap & \name   \\ \hline
  46.86 (5.66)  &  39.27 (2.75) & 52.07 (3.53) & 118.44 (1.98) & 33.85 (6.82) & 33.85 (6.82) & 207.55 (2.72) & \textbf{28.13 (4.06)}  \\
\end{tabular}
\end{adjustbox}
\caption{Mean (and standard deviation) of total running time (in seconds).}
\label{table:online execution time}
\end{table}

For the sake of completion, Appendix \ref{app} contains an instance of the performance of all methods to estimate the true number of clusters.%Finally, let us note that on this toy example, \name\ is the most successful method in estimating the true number $k^\star_t$ of clusters, as shown in .

%\FloatBarrier

\section{Proofs}\label{sec:proofs}
This section contains the proofs to all original results claimed in \hyperref[sec:online clustering method]{Section~\ref*{sec:online clustering method}} %, \autoref{sec:minimax}
and \hyperref[sec:algo]{Section~\ref*{sec:algo}}. 

\subsection{Proof of \autoref{cor:corollary 1}}

Let us first introduce some notation. For any $k\in \llbracket1,p \rrbracket$ and $R>0$, let
\begin{align*}
\mathcal{C}(k,R)&=\left\{\bc=(c_{j})_{j=1,\dots,k}\in\R^{dk}\colon |c_{j}|_{2}\leq R, \forall j\right\},\\
\Xi (k,R)&=\left\{\xi=(\xi_{j})_{j=1,\dots,k}\in\R^{k}\colon 0<\xi_{j}\leq R, \forall j\right\}.
\end{align*}
We denote by $\rho_{k}(\bc,\ttc,\xi)$ the density consisting in the product of $k$ independent uniform distributions on $\ell_{2}$-balls in $\mathbb{R}^{d}$, namely,
\begin{equation*}
\mathrm{d}\rho_{k}(\bc,\ttc,\xi)=\prod_{j=1}^{k}\left\{\frac{\Gamma(\frac{d}{2}+1)}{\pi^{\frac{d}{2}}}\left(\frac{1}{\xi_{j}}\right)^{d}\mathbbm{1}_{\{\mathit{B}_{d}(\ttc_{j},\xi_{j})\}}(c_{j})\right\}\mathrm{d}\bc,
\end{equation*}
where $\ttc \in \mathcal{C}(k,R)$, $\xi \in \Xi (k,R)$ and $\mathit{B}_{d}(\ttc_{j},\xi_{j})$ is an $\ell_{2}$-ball in $\R^{d}$, centered in $\ttc_{j}$ with radius $\xi_{j}$. In the following, we will shorten $\rho_{k}(\bc,\ttc,\xi)$ to $\rho_{k}$ when no confusion can arise. The proof relies on choosing a specific $\rho$ in \autoref{thm:theorem 1}.
For any $k \in \llbracket1,p\rrbracket$, $\ttc \in \mathcal{C}(k,R)$ and $\xi \in \Xi (k,R)$, let $\rho=\rho_{k}\mathbbm{1}_{\{\bc\in\R^{dk}\}}$. Then $\rho$ is a well-defined distribution on $\mathcal{C}$ and belongs to $\PP_\pi(\mathcal{C})$. \autoref{thm:theorem 1} yields
\begin{align}\label{prf:corollary 1}
\expcum\leq&\inf_{k\in \llbracket 1,p\rrbracket}\inf_{\substack{\rho\in \PP_\pi(\mathcal{C})\\\rho=\rho_{k}\mathbbm{1}_{\left\{\bc\in\R^{dk}\right\}}}}\left\{\mathbb{E}_{\bc\sim \rho}\sum_{t=1}^{T}\left[\ell(\bc,x_{t})\right]+\frac{\K(\rho,\pi)}{\lambda}\right. \notag\\ 
&\left.+\frac{\lambda}{2}\mathbb{E}_{(\hat{\rho}_{1},\dots,\hat{\rho}_{T})}\mathbb{E}_{\bc\sim \rho}\sum_{t=1}^{T}\left[\ell(\bc,x_{t})-\ell(\hat{\bc}_{t},x_{t})\right]^2 \right\}.
\end{align}
For any $\rho=\rho_{k}\mathbbm{1}_{\left\{\bc\in\R^{dk}\right\}}$, the first term on the right-hand side of \eqref{prf:corollary 1} satisfies
\begin{align}\label{prf:expected part 1}
\sum_{t=1}^{T}\mathbb{E}_{\bc\sim \rho}\left[\ell(\bc,x_{t})\right]&=
\sum_{t=1}^{T}\mathbb{E}_{\bc\sim \rho_{k}}\left[\ell(\bc,x_{t})\right]\notag\\
&\leq\sum_{t=1}^{T}\min_{j=1,\dots,k}\left\{\mathbb{E}_{\bc \sim \rho_{k}}\left[|c_{j}-\ttc_{j}|_{2}^{2}\right]+|\ttc_{j}-x_{t}|_{2}^{2}\right\}\notag\\
&=\sum_{t=1}^{T}\min_{j=1,\dots,k}\left\{\frac{d}{d+2}\xi_{j}^{2}+|\ttc_{j}-x_{t}|_{2}^{2}\right\}\notag\\
&\leq \frac{dT}{d+2}\max_{j=1,\dots,k}\xi_{j}^{2}+\sum_{t=1}^{T}\ell(\ttc,x_{t}).
\end{align}
Let us now compute the second term on the right-hand side of \eqref{prf:corollary 1}.
\begin{align*}
\K(\rho,\pi)&=\int_{\mathcal{C}}\log\frac{\rho(\bc)}{\pi(\bc)}\rho(\bc) \mathrm{d}\bc \\ &=\int_{\R^{dk}}\left(\log\frac{\rho_{k}(\bc)}{\pi_{k}(\bc)}+\log\frac{\pi_{k}(\bc)}{\pi(\bc)}\right)\rho_{k}(\bc)\mathrm{d}\bc\notag\\
&=\K(\rho_{k},\pi_{k})+\log\frac{1}{q(k)}\\ &=:A+B,
\end{align*}
where
\begin{align*}
A=\int_{\R^{dk}} \log \prod_{j=1}^{k}\frac{\left(\frac{1}{\xi_{j}}\right)^{d}}{\left(\frac{1}{2R}\right)^{d}}\rho_{k}(\bc)\mathrm{d}\bc=d\sum_{j=1}^{k}\log \left(\frac{2R}{\xi_{j}}\right).
\end{align*}
Since the function $x \mapsto (1-e^{-\eta x})/x$ is non-increasing for $x>0$ and $\eta>0$, we have
\begin{align}\label{ieq:log discrete probability}
B&=\log \left(\frac{e^{-\eta}(1-e^{-\eta p})}{1-e^{-\eta}}e^{\eta k}\right)\notag\\
&\leq \log\left(pe^{\eta(k-1)}\right)\notag \\ &=\eta(k-1)+\log p.
\end{align}
When $\eta=0$, $q$ is a uniform distribution on $\llbracket 1,p\rrbracket$, and the above inequality holds as well.
Then, $\K(\rho,\pi)/\lambda$ in \eqref{prf:corollary 1} may be upper bounded as follows:
\begin{equation}\label{prf:second term}
\frac{\K(\rho,\pi)}{\lambda}\leq \frac{d}{\lambda}\sum_{j=1}^{k}\log \left(\frac{2R}{\xi_{j}}\right)+\frac{\eta(k-1)}{\lambda}+\frac{\log p}{\lambda}.
\end{equation}
Finally,
\begin{align*}
\left|\ell(\bc,x_{t})-\ell(\hat{\bc}_{t},x_{t})\right|&=\left|\min_{j=1,\dots,k}|c_{j}-x_{t}|_{2}^{2}-\min_{j=1,\dots,K_{t}}|\hat{c}_{t,j}-x_{t}|_{2}^{2}\right|\notag\\
&\leq \left(2R+\max_{t=1,\dots,T}|x_{t}|_{2}\right)^{2} =:C_{1}.
\end{align*}
Then, the third term of the right-hand side in \eqref{prf:corollary 1} is controlled as
\begin{align}\label{prf:third term}
\frac{\lambda}{2}\mathbb{E}_{(\hat{\rho}_{1},\dots,\hat{\rho}_{T})}\mathbb{E}_{\bc\sim \rho_{k}}\sum_{t=1}^{T}\left[\ell(\bc,x_{t})-\ell(\hat{\bc}_{t},x_{t})\right]^2
\leq \frac{\lambda T}{2}C_{1}^{2}.
\end{align}
Combining inequalities \eqref{prf:expected part 1}, \eqref{prf:second term} and \eqref{prf:third term} gives, for any $\xi \in \Xi (k,R)$,
\begin{align*}
\expcum\leq&\inf_{k\in \llbracket 1,p\rrbracket}\inf_{\ttc\in\mathcal{C}(k,R)}\left\{\sum_{t=1}^{T}\ell(\ttc,x_{t})+\frac{dT}{d+2}\max_{j=1,\dots,k}\xi_{j}^{2}\right. \notag\\
&\left.+\frac{d}{\lambda}\sum_{j=1}^{k}\log \left(\frac{2R}{\xi_{j}}\right)+\frac{\eta}{\lambda}(k-1)\right\}+\frac{\lambda T}{2}C_{1}^{2}+\frac{\log p}{\lambda}.
\end{align*}
Under the assumption that $\lambda>(d+2)/(2TR^{2})$, the global minimizer of the function 
\begin{align}\label{eq:minimizer function}
(\xi_1,\dots,\xi_k)\mapsto \frac{Td}{d+2}\max_{j=1,\dots,k}\xi_{j}^{2}+\frac{d}{\lambda}\sum_{j=1}^{k}\log \left(\frac{2R}{\xi_{j}}\right)
\end{align}
does not necessarily belong to $\Xi (k,R)$.  
A possible choice of $(\xi_{j})_{1:k}\in \Xi (k,R)$ is given by
\begin{equation*}
\xi^{\star}_{1}=\xi^{\star}_{2}=\dots=\xi^{\star}_{k}=\sqrt{\frac{d+2}{2\lambda T}}.
\end{equation*}
Then \eqref{eq:minimizer function} amounts to
\begin{align*}
\frac{d}{2\lambda}+\frac{dk}{2\lambda}\log\left(\frac{8R^{2}\lambda T}{d+2}\right).
\end{align*}
Hence,
\begin{align*}
\expcum\leq&\inf_{k\in \llbracket 1,p\rrbracket}\inf_{\ttc\in\mathcal{C}(k,R)}\left\{\sum_{t=1}^{T}\ell(\ttc,x_{t})+\frac{dk}{2\lambda}\log\left(\frac{8R^{2}\lambda T}{(d+2)k}\right) +\frac{\eta}{\lambda}k\right\} \\ &+\left(\frac{\log p}{\lambda}+\frac{d}{2\lambda}+\frac{\lambda T}{2}C_{1}^{2}\right).
\end{align*}

\subsection{Proof of \autoref{thm:theorem 2}}

The proof builds upon the online variance inequality described in \cite{Aud2009}, \emph{i.e.}, for any $\lambda>0$, any $\hat{\rho} \in \PP_\pi(\mathcal{C})$ and any $x\in \R^{d}$,
\begin{equation}\label{eq:online variance inequality}
\mathbb{E}_{\bc^{\prime}\sim\hat{\rho}} [\ell(\bc^{\prime},x)]\leq -\frac{1}{\lambda}\mathbb{E}_{\bc^{\prime}\sim \hat{\rho}} \log \mathbb{E}_{\bc\sim \hat{\rho}}\left[e^{-\lambda\left[\ell(\bc,x)+\frac{\lambda}{2}\left(\ell(\bc,x)-\ell(\bc^{\prime},x)\right)^{2}\right]}\right].
\end{equation}
By \eqref{eq:online variance inequality}, we have
\begin{align}\label{eq:basic inequality}
\expcum\notag=&\sum_{t=1}^{T}\mathbb{E}_{(\hat{\rho_{1}},\dots,\hat{\rho}_{t-1})}\mathbb{E}_{\hat{\rho}_{t}}\left[\ell(\hat{\bc}_{t},x_{t})\mid\hat{\bc}_{1},\dots,\hat{\bc}_{t-1}\right]\notag\\
\leq&\sum_{t=1}^{T}\mathbb{E}_{(\hat{\rho_{1}},\dots,\hat{\rho}_{t-1})}\left[-\frac{1}{\lambda_{t-1}}\mathbb{E}_{\hat{\bc}_{t}\sim \hat{\rho}_{t}}\log \mathbb{E}_{\bc\sim \hat{\rho}_{t}}\left(e^{-\lambda_{t-1}[\ell(\bc,x_{t})+\frac{\lambda_{t-1}}{2}(\ell(\bc,x_{t})-\ell(\hat{\bc}_{t},x_{t}))^{2}]}\right)\right]\notag\\
\leq& \mathbb{E}_{(\hat{\rho_{1}},\dots,\hat{\rho}_{T})}\left[\sum_{t=1}^{T}-\frac{1}{\lambda_{t-1}}\log \frac{\int e^{-\lambda_{t-1}S_{t}(\bc)\mathrm{d}\pi(\bc)}}{\int e^{-\lambda_{t-1}S_{t-1}(\bc)\mathrm{d}\pi(\bc)}}\right]\notag \\ =&\mathbb{E}_{(\hat{\rho_{1}},\dots,\hat{\rho}_{T})}\left[\sum_{t=1}^{T}-\frac{1}{\lambda_{t-1}}\log \frac{V_{t}}{W_{t-1}}\right]\notag\\
=&\mathbb{E}_{(\hat{\rho_{1}},\dots,\hat{\rho}_{T})}\left[\sum_{t=1}^{T}\left[\frac{1}{\lambda_{t-1}}\log W_{t-1}-\frac{1}{\lambda_{t-1}}\log V_{t}\right]\right].
\end{align}
Applying Jensen's inequality, for any $ 1\leq t\leq T$,
\begin{align*}
\frac{1}{\lambda_{t-1}}\log V_{t}&=\frac{1}{\lambda_{t-1}}\log \mathbb{E}_{\bc\sim \pi}\left[\left(e^{-\lambda_{t}S_{t}(\bc)}\right)^{\frac{\lambda_{t-1}}{\lambda_{t}}}\right]\notag\\
&\geq \frac{1}{\lambda_{t-1}}\log \left(\mathbb{E}_{\bc\sim \pi}\left[e^{-\lambda_{t}S_{t}(\bc)}\right]\right)^{\frac{\lambda_{t-1}}{\lambda_{t}}}\\ &=\frac{1}{\lambda_{t}}\log W_{t}.
\end{align*}
Therefore, since $W_{0}=1$,
\begin{align}\label{eq:jensen in adaptive case}
\sum_{t=1}^{T}\left[\frac{1}{\lambda_{t-1}}\log W_{t-1}-\frac{1}{\lambda_{t-1}}\log V_{t}\right]\leq -\frac{1}{\lambda_{T}}\log W_{T},
\end{align}
and by \eqref{eq:basic inequality}, \eqref{eq:jensen in adaptive case} and the duality formula \eqref{eq:duality formula}, we have
\begin{align*}
\expcum
\leq& \mathbb{E}_{(\hat{\rho_{1}},\dots,\hat{\rho}_{T})}\left[-\frac{1}{\lambda_{T}}\log \mathbb{E}_{\bc\sim \pi}\left[e^{-\lambda_{T}S_{T}(\bc)}\right]\right]\\ \leq& -\frac{1}{\lambda_{T}}\log \mathbb{E}_{\bc\sim \pi}\left[e^{-\lambda_{T}\mathbb{E}_{(\hat{\rho_{1}},\dots,\hat{\rho}_{T})}S_{T}(\bc)}\right]\qquad\textrm{\citep[by][Lemma 3.2]{Aud2009}}\\
=&\inf_{\rho \in \PP_\pi(\mathcal{C})}\left\{\mathbb{E}_{\bc\sim \rho}\left[\sum_{t=1}^{T}\ell(\bc,x_{t})\right]+\mathbb{E}_{\bc\sim \rho}\mathbb{E}_{(\hat{\rho_{1}},\dots,\hat{\rho}_{T})}\left[\sum_{t=1}^{T}\frac{\lambda_{t-1}}{2}\left(\ell(\bc,x_{t})-\ell(\hat{\bc}_{t},x_{t})\right)^{2}\right]\right.\\
&\left.+\frac{\K(\rho,\pi)}{\lambda_{T}}\right\},
\end{align*}
which achieves the proof.

\subsection{Proof of \autoref{cor:corollary 3}}

The proof is similar to the proof of \autoref{cor:corollary 1},  the only difference lies in the fact that \eqref{prf:third term} is replaced with
\begin{align*}
\mathbb{E}_{(\hat{\rho}_{1},\dots,\hat{\rho}_{T})}\mathbb{E}_{\bc\sim \rho_{k}}\sum_{t=1}^{T}\frac{\lambda_{t-1}}{2}[\ell(\bc,x_{t})-\ell(\hat{\bc}_{t},x_{t})]^2
&\leq \frac{(d+2)C_{1}^{2}}{4R^{2}}\left(1+\sum_{t=2}^{T}\frac{\sqrt{\log (t-1)}}{\sqrt{t-1}}\right) \\
&\leq \frac{(d+2)C_{1}^{2}}{4R^{2}}\left(1+\frac{\sqrt{\log 2}}{\sqrt{2}} +\frac{\sqrt{\log 3}}{\sqrt{3}} + \sum_{t=4}^{T-1}\int_{t-1}^{t}\frac{\sqrt{\log x}}{\sqrt{x}}\mathrm{d}x \right)\\
&\leq \frac{(d+2)C_{1}^{2}}{2R^{2}}\sqrt{T\log T},
\end{align*}
where the second inequality above is due to the fact that $\frac{\sqrt{\log t}}{\sqrt{t}} \leq \int_{t-1}^{t}\frac{\sqrt{\log x}}{\sqrt{x}}\mathrm{d}x$ when $t\geq 4$ and the last inequality is deduced from the following with change of variable $y = \sqrt{\log x}$, \emph{i.e.},
\begin{align*}
    \int_{3}^{T-1}\frac{\sqrt{\log x}}{\sqrt{x}}\mathrm{d}x & = \int_{\sqrt{\log 3}}^{\sqrt{\log (T-1)}}2y^2 e^{\frac{y^2}{2}}\mathrm{d}y\\
    &\leq \sqrt{\log (T-1)}\int_{\sqrt{\log 3}}^{\sqrt{\log (T-1)}}2y e^{\frac{y^2}{2}}\mathrm{d}y\\
    & = 2\sqrt{\log (T-1)}\left(\sqrt{T-1}-\sqrt{3}\right).
\end{align*}

\subsection{Proof of \autoref{cor:corollary 4}}

Let us denote by $M$ the index of the last epoch and let $t_{M} = T$. We assume $M \geq 1$ (otherwise, the corollary follows directly from \autoref{cor:corollary 3} applied with an upper bound $R_{0}$ of $\ell_{2}$-norm of sequence $(x_{t})_{1:T}$). If $R_{t_{M}}\leq R_{t_{M-1}}$, then we have $R_{T} = R_{t_{M}} = R_{t_{M-1}}$, hence one always has $R_{t_{M}} \geq R_{t_{M-1}}$. In addition, since $M \geq 1$, we also have $R_{t_{M}} \leq 2\max_{t=1,\dots,T}\left|x_{t}\right|_{2} = 2R$. 
\medskip

Let us introduce for each epoch $r, r=0,1,\dots, M$ the following notation

\begin{equation*}
    E^{(r)} = \sum_{t= t_{r-1}+1}^{t_{r}-1}\E_{\left(\hat{\rho}_{1}, \dots, \hat{\rho}_{t}\right)}\ell\left(\hat{\bc}_{t}, x_{t}\right),
\end{equation*}
and for $k \in \llbracket1,p\rrbracket$ , $\bc \in \mathcal{C}(k, R)$

\begin{equation*}
    L^{(r)}(k, \bc) = \sum_{t= t_{r-1}+1}^{t_{r}-1}\ell(\bc, x_{t}).
\end{equation*}

Within each epoch $r=0,1,\dots,M$, since 
\begin{equation}\label{ineq:epoch bound}
  \max_{t=t_{r-1}+1, t_{r-1}+2, \dots, t_{r}-1}|x_{s}|_{2} \leq R_{t_{r-1}},
\end{equation}
then applying \autoref{cor:corollary 3} to each epoch $r$ can give us that, for each $k \in \llbracket1, p\rrbracket$, 

\begin{equation}\label{ineq:corollary 3 on epoch}
    E^{(r)} - \inf_{\bc \in \mathcal{C}\left(k, R_{t_{r-1}}\right)}L^{(r)}(k,\bc)  \leq \left(C(d, \eta)k+C(p, d)\right) R^{2}_{t_{r-1}}\sqrt{(t_{r}-1)\log (t_{r}-1)},
\end{equation}
where $C(d,\eta) = \frac{2(d+\eta)}{d+2}$ and $C(p,d) = \frac{2\log p + d}{d+2} +\frac{81(d+2)}{2}$.
\medskip

In addition, since all observations $x_{t}, t= t_{r-1}+1, , \dots, t_{r}-1$ in the epoch $r$ are bounded in a convex ball $\mathit{B}_{d}\left( R_{t_{r-1}}\right)$, centered in $0 \in \R^{d}$ with radius  $R_{t_{r-1}}$ as indicated by \eqref{ineq:epoch bound}, we have for each $\bc^{\prime} \in \mathcal{C}(k, R)\setminus \mathcal{C}\left(k, R_{t_{r-1}}\right)$, $k=1,2,\dots,p$ that 

\begin{equation}\label{ineq:convex infimum}
    \inf_{\bc \in \mathcal{C}\left(k, R_{t_{r-1}}\right)}L^{(r)}(k, \bc) \leq L^{(r)}(k, \bc^{\prime}).
\end{equation}
By \eqref{ineq:corollary 3 on epoch} and \eqref{ineq:convex infimum}, we can have that for any $k \in \llbracket1, p\rrbracket$ and $\bc \in \mathcal{C}(k, R)$, the following inequality holds, 
\begin{equation*}
    E^{(r)} -  L^{(r)}(k, \bc) \leq \left(C(d, \eta)k+C(p, d)\right) R^{2}_{t_{r-1}}\sqrt{(t_{r}-1)\log (t_{r}-1)}.
\end{equation*}
Therefore, for any $\bc \in \mathcal{C}(k, R)$, one has
\begin{align*}
    \sum_{t= 1}^{T}\E_{\left(\hat{\rho}_{1}, \dots, \hat{\rho}_{t}\right)}\ell\left(\hat{\bc}_{t}, x_{t}\right) - \sum_{t=1}^{T}\ell(\bc, x_{t}) &= \sum_{r=0}^{M}\left(E^{(r)} - L^{(r)}(k, \bc)\right) + \sum_{r=0}^{M}\left(\E_{\left(\hat{\rho}_{1},\dots, \hat{\rho}_{t_{r}}\right)}\ell\left(\hat{\bc}_{t_{r}}, x_{t_{r}}\right) - \ell(\bc, x_{t_{r}})\right) \\
    &\leq \sum_{r=0}^{M}\left[C(d,\eta)k + C(p, d)\right]R_{t_{r-1}}^{2}\sqrt{(t_{r}-1)\log (t_{r}-1)} + 4\sum_{r=0}^{M}R_{t_{r}}^{2}\\
    &\leq \sum_{r=0}^{M}\left[C(d,\eta)k + C(p, d)\right]R_{t_{r-1}}^{2}\sqrt{T\log T} + 4\sum_{r=0}^{M}R_{t_{r}}^{2}.
\end{align*}
Since $R_{t_{s}}\geq 2^{s-r}R_{t_{r}}$ for $0\leq r \leq s \leq M-1$, then for $s\leq M-1$, 
\begin{equation*}
    \sum_{r=0}^{s}R^{2}_{t_{r}} \leq \sum_{r=0}^{s}4^{r-s}R^{2}_{t_{s}}\leq \frac{4}{3}R^{2}_{t_{s}}.
\end{equation*}
Hence,
\begin{align*}
\sum_{r=0}^{M}R^{2}_{t_{r-1}} &\leq R_{t_{-1}}^{2} + \frac{4}{3}R_{t_{M-1}}^{2}\leq \frac{7}{3}R_{t_{M}}^{2} \\
4\sum_{r=0}^{M}R^{2}_{t_{r}} &\leq 4\left(\frac{4}{3}R_{t_{M-1}}^{2} +R_{t_{M}}^{2} +  \right)\leq \frac{28}{3}R_{t_{M}}^{2} \\
\end{align*}
Therefore,
\begin{align*}
\sum_{t= 1}^{T}\E_{\left(\hat{\rho}_{1}, \dots, \hat{\rho}_{t}\right)}\ell\left(\hat{\bc}_{t}, x_{t}\right) - \sum_{t=1}^{T}\ell(\bc, x_{t}) &\leq \frac{7}{3}\left[C(d,\eta)k + C(p, d)\right]R_{t_{M}}^{2}\sqrt{T\log T} + \frac{28}{3}R_{t_{M}}^{2} \\
&\leq \frac{28}{3}\left[C(d,\eta)k + C(p, d)\right]R^{2}\sqrt{T\log T} + \frac{112}{3}R^{2},
\end{align*}
where $R = \max_{t=1,2,\dots, T}|x_{t}|_{2}$ and the second inequality is due to the fact that $R_{t_{M}}\leq 2R$. 
Taking the infimum of $\sum_{t=1}^{T}\ell(\bc, x_{t})$ over the set $\mathcal{C}(k, R)$, $k \in \llbracket 1, p \rrbracket$ leads to 
\begin{equation*}
\sum_{t= 1}^{T}\E_{\left(\hat{\rho}_{1}, \dots, \hat{\rho}_{t}\right)}\ell\left(\hat{\bc}_{t}, x_{t}\right) \leq \inf_{\bc \in \mathcal{C}(k, R)}\sum_{t=1}^{T}\ell(\bc, x_{t}) + \frac{28}{3}\left[C(d,\eta)k + C(p, d)\right]R^{2}\sqrt{T\log T} + \frac{112}{3}R^{2}.
\end{equation*}
Finally, taking the infimum of the right hand side of the above inequality with respect to $k$ terminates the proof.

\subsection{Proof of \autoref{thm:theorem 3}}

The proof for the upper bound is straightforward: by replacing the loss function $\ell(\bc,x)$ by the penalized loss $\ell_\alpha(\bc,x)=\ell(\bc,x)+\alpha|\bc|$ with $\alpha=\sqrt{\log T}/\sqrt{T}$ in the proof of \autoref{thm:theorem 2}, we obtain
\begin{multline*}
\sum_{t=1}^{T}\mathbb{E}_{\hat{\rho}_{1},\dots,\hat{\rho_{t}}}\ell_{\alpha}(\hat{\bc}_t,x_t) \leq \inf_{\rho \in \PP_\pi(\mathcal{C})}\left\{\mathbb{E}_{\bc\sim \rho}\left[\sum_{t=1}^{T}\ell_{\alpha}(\bc,x_{t})\right]+\frac{\K(\rho,\pi)}{\lambda_{T}} \right. \\ \left.
+\mathbb{E}_{(\hat{\rho}_{1},\dots,\hat{\rho}_{T})}\mathbb{E}_{\bc\sim \rho}\left[\sum_{t=1}^{T}\frac{\lambda_{t-1}}{2}[\ell_{\alpha}(\bc,x_{t})-\ell_{\alpha}(\hat{\bc}_{t},x_{t})]^2\right]\right\},
\end{multline*}
and choosing $\lambda=\sqrt{\log T}/\sqrt{T}$ and $p=T^{\frac{1}{4}}$ yields the desired upper bound.
\medskip

We now proceed to the proof of the lower bound. The trick is to replace the supremum over the $(x_{t})$ in $\mathcal{V}_T(s)$ by an expectation.
\medskip

We first introduce the event $\Omega_{s,R}=\left\{\left(X_1, \ldots, X_T\right) \in \R^{dT}: \ \text{such that} \ \left|\bc^{\star}_{T,R}\right|=s \right\}$, where $\bc^\star_{T,R}$ is defined as in \textbf{Assumption} $\mathcal{H}(s)$. Then, we have
\begin{align*}
\mathcal{V}_T(s) 
\geq \inf_{(\hat{\rho}_{t})}\ \E_{\mu^{ T}}\left\{\sum_{t=1}^T \E_{(\hat{\rho}_{1}, \dots, \hat{\rho}_{t})}\left(\ell(\hat\bc_t,X_t)+\frac{\sqrt{\log T}}{\sqrt{T}} |\hat\bc_t|\right)-
\infpart\right\}\mathbbm{1}\left(\Omega_{s,R}\right),
\end{align*}
where $\mu^{T} \in \PP(\R^{dT})$ is the joint distribution of i.i.d. sample $(X_1, \ldots, X_T)$. Now, we have to choose $\mu$ in order to maximize the right-hand side of the above inequality. This is the purpose of the following lemmas.
\begin{lem}
\label{lemma:optimal partitions}
Let $s\in \mathbb{N}^{*}, s\leq p$. Let $\mu\in \PP (\R^d)$ a distribution concentrated on $2s$ fixed points $\mathcal{S}_\mu =\{z_i,z_i+w, i=1, \dots, s\}$ such that $w=(2\Delta, 0, \dots, 0)\in\R^d$ with $\Delta>0$ and that $z_{1},\dots,z_{s}\in \mathit{B}_{d}(R)$. Suppose that for any $i\not=j$,  $d(z_i,z_j)\geq 2A\Delta$ for some $A>0$. Define $\mu$ as the uniform distribution over 
$\mathcal{S}_\mu$. Then, if $A > \sqrt{2}+1$, we have
\begin{equation*}
\arg\inf_{\bc\in\CC(s,R)}\E_\mu\ell(\bc,X)=\{z_{i}+w/2, \quad i =1,\dots, s\}=:\bc^{\star}_{\mu,s}.
\end{equation*}
\end{lem}
The proof of \autoref{lemma:optimal partitions} is similar to \citet[][Section III.A, step 3]{BLL1998}.
The next lemma controls the probability of the event $|\bc_{T, R}^{\star}| \neq s $ with a proper choice of $\Delta^2$ and $A$ in the definition of $\mu$.
\begin{lem}
\label{lemma:choice of delta}
Let $s\in \mathbb{N}^*$, $2\leq s\leq p$, and $\mu$ is defined in \autoref{lemma:optimal partitions}. Then, if we choose $A = \sqrt{2}s + 1$ and
\begin{equation*}
\frac{2(s-1)s\sqrt{\log T}}{(A-1)^2\sqrt{T}} < \Delta^2 < \frac{\sqrt{\log T}}{\sqrt{T}},    
\end{equation*}
then for any $\epsilon > 0$ and $T > 8s^2\log\frac{2s^2}{\epsilon}$, we have
\begin{equation*}
\mathbb{P}\left(\left|\bc^\star_{T, R}\right| \neq s \right) \leq \epsilon.    
\end{equation*}
%$$
%where $(X_1, \ldots, X_T)$ are i.i.d. with law $\mu$ and:
%$$
%\bc^\star_T:=\arg\min_{\bc\in\R^{dT}}\left\{\sum_{t=1}^T\ell(\bc,X_t)+\sqrt{T}\log T |\bc|_0\right\}.
%$$
\end{lem}
\begin{proof} For any $k \in \llbracket1, p\rrbracket$,  let $\bc^{\star}_{T,k}$ firstly denote the optimal partition in $\mathcal{C}(k, R)$ that minimizes the penalized empirical loss on $(X_{1}, \dots, X_{T})$, \emph{i.e.},  
\begin{equation*}
\bc_{T, k}^{\star} = \arg\inf_{\bc\in\CC(k,R)}\left\{\frac{1}{T}\sum_{t=1}^{T}\ell(\bc, X_{t}) + \left|\bc\right|\frac{\sqrt{\log T}}{\sqrt{T}}\right\}.   
\end{equation*}
In addition, denote by $\bc^{\star}_{\mu, k}$ the partition minimizing the expected penalized loss, \emph{i.e.},
\begin{equation*}
\bc_{\mu, k}^{\star} = \arg\inf_{\bc\in\CC(k,R)}\left\{\E_{\mu}\ell(\bc, X) + \left|\bc\right|\frac{\sqrt{\log T}}{\sqrt{T}}\right\}.   
\end{equation*}
One can notice that in fact $|\bc| = k$ in the two above definitions for any $\bc \in\CC(k,R) \in \R^{dk}$.
Next
\begin{align}\label{ieq:bigger than s}
\mathbb{P}\left(\left|\bc^{\star}_{T,R}\right| > s\right) &=\sum_{k=s+1}^{2s}\mathbb{P}\left(\left|\bc^{\star}_{T,R}\right| = k\right)  \notag \\ 
& \leq \sum_{k=s+1}^{2s}\mathbb{P}\left(\frac{1}{T} \sum_{t=1}^{T}\ell\left(\bc^{\star}_{T,k-1}, X_{t}\right) - \frac{1}{T} \sum_{t=1}^{T}\ell\left(\bc^{\star}_{T,k}, X_{t}\right) > \sqrt{\frac{\log T}{T}} \right) \notag \\
&\leq \sum_{k=s+1}^{2s}\mathbb{P}\left(\frac{1}{T}\sum_{t=1}^{T}\ell\left(\bc^{\star}_{T,k-1}, X_{t}\right) > \sqrt{\frac{\log T}{T}}\right) \notag \\
&\leq s\mathbb{P}\left(\frac{1}{T}\sum_{t=1}^{T}\ell\left(\bc^{\star}_{\mu, s}, X_{t}\right) > \sqrt{\frac{\log T}{T}}\right) \notag \\
&=s\mathbb{P}\left(\Delta^{2} > \sqrt{\frac{\log T}{T}}\right) = 0,
\end{align}
where the first inequality is induced by the definition of $\bc_{T, R}^{\star}$ and the third inequality is due to the fact that we have almost surely
\begin{equation*}
\sum_{t=1}^{T}\ell\left(\bc_{\mu,s}^{\star}, X_{t}\right) \geq \sum_{t=1}^{T}\ell\left(\bc_{T,s}^{\star}, X_{t}\right) \geq \sum_{t=1}^{T}\ell\left(\bc_{T,k-1}^{\star}, X_{t}\right),\quad \text{for } k > s.
\end{equation*}
In order to control the probability  $\mathbb{P}(|\bc^{\star}_{T,R}| < s)$, let us first consider the Voronoi partition of $\R^{d}$ induced by the set of points $\{z_{i}, z_{i} + w, i = 1,\dots,s\}$ and for each $i$ define $V_{i}$ as the union of the Voronoi cells belonging to $z_{i}$ and $z_{i}+w$. Let $N_{i}$ denotes the number of $X_{t}$, $t=1,\dots, T$  falling in $V_{i}$. Hence $(N_{1}, \dots, N_{s})$ follows a multinomial distribution with parameter $(T, q_{1}, q_{2}, \dots, q_{s})$, where $q_{1} = q_{2} = \cdots = q_{s} = 1/s$. Then
\begin{align}
\mathbb{P}\left(\left|\bc^{\star}_{T,R}\right| < s\right) &=\sum_{k=1}^{s-1}\mathbb{P}\left(\left|\bc^{\star}_{T,R}\right| = k\right) \notag\\
&\leq \sum_{k=1}^{s-1}\mathbb{P}\left(\frac{1}{T}\sum_{t=1}^{T}\ell\left(\bc^{\star}_{T,k}, X_{t}\right) - \frac{1}{T}\sum_{t=1}^{T}\ell\left(\bc_{T,s}^{\star},X_{t}\right) \leq \frac{(s-k)\sqrt{\log T}}{\sqrt{T}}\right) \notag \\
&\leq \sum_{k=1}^{s-1}\mathbb{P}\left(\frac{1}{T}\sum_{t=1}^{T}\ell\left(\bc^{\star}_{T,k}, X_{t}\right) - \frac{1}{T}\sum_{t=1}^{T}\ell\left(\bc_{\mu,s}^{\star},X_{t}\right) \leq \frac{(s-k)\sqrt{\log T}}{\sqrt{T}}\right) \notag \\
&\leq (s-1)\mathbb{P}\left(\frac{1}{T}\min_{i=1,\dots, s}N_{i}\cdot (A-1)^2\Delta^2 - \Delta^{2} \leq \frac{(s-k)\sqrt{\log T}}{\sqrt{T}}\right) \notag \\
&\leq(s-1)s\mathbb{P}\left(N_{1} \leq \frac{T\Delta^{2}+(s-1)\sqrt{T\log T}}{(A-1)^2\Delta^{2}}\right). \notag 
\end{align}
The third inequality is due to the fact that $\sum_{t=1}^{T}\ell\left(\bc^{\star}_{T,k}, X_{t}\right) \geq \min_{i=1,\dots, s}N_{i} (A-1)^2\Delta^2$ for $k<s$, and the last inequality holds since the marginal distributions of the $N_{i}$s ($i=1,\dots, s$)  are the same binomial distribution with parameter $(T, 1/s)$. Finally, we can bound the last term by Hoeffding's inequality, \emph{i.e.,} for any $t>0$
\begin{equation*}
\mathbb{P}\left(N_{1} - \E\left(N_{1}\right) \leq -t\right) \leq 2\exp\left(-\frac{2t^{2}}{T}\right).
\end{equation*}
Hoeffding's inequality implies that if  $s>2, A = \sqrt{2}s+1, T> 8s^2 \log \frac{2s^2}{\epsilon}$ and $\Delta^2 > \frac{2s(s-1)\sqrt{\log T}}{(A-1)^2\sqrt{T}}$, then 
\begin{equation*}
\mathbb{P}\left(N_{1} \leq \frac{T\Delta^{2}+(s-1)\sqrt{T \log T}}{(A-1)^2\Delta^{2}}\right) < \frac{\epsilon}{s^2}.
\end{equation*}
\end{proof}
%Hence we terminate the proof of \autoref{lemma:choice of delta}.
Next, we proceed to the proof of \autoref{thm:theorem 3}. 
%$\nu^{N}\in \PP(\R^{dsN})$ is the distribution of an i.i.d. sample $(C_1, \ldots, C_N)$ of partition such that for any $j$, $C_j \in \CC(s,R)$ $\nu$-a.s. 
First of all, since $(X_1, \ldots, X_T)$ are i.i.d, following the distribution $\mu$ and by the definition of $\Omega_{s,R}$, we can write
\begin{align*}
\inf_{(\hat{\rho}_{t})}\ &\E_{\mu^{ T}}\left\{\sum_{t=1}^T \E_{(\hat{\rho}_{1}, \dots, \hat{\rho}_{t})}\left(\ell(\hat\bc_t,X_t)+\sqrt{\frac{\log T}{T}} |\hat\bc_t|\right)\right\}\mathbbm{1}\left(\Omega_{s,R}\right) \\
= & \inf_{(\hat{\rho}_{t})}\ \E_{(\hat{\rho}_{1}, \dots,\hat{\rho}_{T})}\sum_{t=1}^{T}\E_{\mu^T}\left[\left(\ell(\hat{\bc}_{t}, X_{t}) + \sqrt{\frac{\log T}{T}}|\hat{\bc}_{t}|\right)\mathbbm{1}(\Omega_{s,R})\right] \\
\geq &\inf_{\hat{\bc}}\ \E_{\mu^{T}}\left\{\sum_{t=1}^T\ell(\hat{\bc},X_t)+\sqrt{T \log T} |\hat{\bc}|\right\}\mathbbm{1}\left(\Omega_{s,R}\right)\\
\geq &\E_{\mu^{T}}\left\{\sum_{t=1}^T\ell(\bc^{\star}_{T,R},X_t)+s\sqrt{T \log T} \right\}\mathbbm{1}\left(\Omega_{s,R}\right)\\
\geq &\E_{\mu^{T}}\left\{\sum_{t=1}^T\ell(\bc^{\star}_{T,R},X_t)\right\}\left(1-\mathbbm{1}\left(\Omega^C_{s,R}\right)\right)+s\sqrt{T\log T}\mathbb{P}\left(\Omega_{s,R}\right) \\
\geq &\E_{\mu^{T}}\left\{\sum_{t=1}^T\ell(\bc^{\star}_{T,R},X_t)\right\}-T\Delta^2\mathbb{P}\left(\Omega_{s,R}^C\right) +s\sqrt{T \log T }\left(\mathbb{P}\left(\Omega_{s,R}\right)-\mathbb{P}\left(\Omega_{s,R}^{C}\right)\right)\\
\geq & T\inf_{\bc \in \mathcal{C}(s,R)}\E_\mu\ell(\bc,X)-T\Delta^{2} \mathbb{P}\left(\Omega_{s,R}^C\right)+s\sqrt{T \log T}\left(\mathbb{P}\left(\Omega_{s,R}\right)-\mathbb{P}\left(\Omega_{s,R}^{C}\right)\right),
\end{align*}
where $\hat{\bc}$ in the first inequality is given by $$\hat{\bc} = \arg\inf_{\bc\in \CC}\E_{\mu^T}\left[\left(\ell(\bc,X_{t})+|\bc|\sqrt{\log T}/\sqrt{T}\right)\mathbb{1}(\Omega_{s,R})\right].$$ Note that $\hat{\bc}$ does not depend on $t$ since $\mu$ is a symmetric uniform distribution (definition in \autoref{lemma:optimal partitions}).
The second inequality is due to Jensen's inequality and the fourth inequality relies on the fact that with the definition of $\bc_{T,R}^{\star}$ and $\mu$, we have almost surely that
\begin{equation*}
\sum_{t=1}^{T}\ell\left(\bc_{T,R}^{\star}, X_{t}\right) \leq \sum_{t=1}^{T}\ell\left(\bc_{\mu,s}^{\star}, X_{t}\right) + s\sqrt{T \log T} = T\Delta^2 + s\sqrt{T \log T},    
\end{equation*}
where $\Delta>0$ is related with the choice of $\mu$ in \autoref{lemma:optimal partitions} and its value is constrained according to \autoref{lemma:choice of delta}. Then we obtain for any $\epsilon >0$
\begin{multline}\label{ieq:minimax part 1}
\inf_{(\hat{\rho}_{t})}\ \E_{\mu^{ T}}\left\{\sum_{t=1}^T\E_{(\hat{\rho}_{1}, \dots, \hat{\rho}_{t})}\ell(\hat\bc_t,X_t)+\frac{\sqrt{\log T}}{\sqrt{T}} |\hat\bc_t|\right\}\mathbbm{1}\left(\Omega_{s,R}\right)\geq T\inf_{\bc \in \mathcal{C}(s,R)}\E_\mu\ell(\bc,X)-T\epsilon\Delta^2 \\ +s\sqrt{T\log T}(1-2\epsilon).
\end{multline}
Moreover, by Jensen's inequality
\begin{align}\label{ieq:minimax part 2}
\E_{\mu^T}\left[\inf_{\bc \in \mathcal{C}(s,R)}\sum_{t=1}^{T}\ell\left(\bc, X_{t}\right)\mathbb{1}\left(\Omega_{s,R}\right)\right] \leq T\inf_{\bc \in \mathcal{C}(s, R)}\E_{\mu}\ell(\bc, X).
\end{align}
Combining \eqref{ieq:minimax part 1} and \eqref{ieq:minimax part 2}, we obtain
\begin{align}
\label{denguin}
\mathcal{V}_T(s)\geq s\sqrt{T \log T} \left(1-2\epsilon\left[1+\frac{\sqrt{T}\Delta^2}{2s\sqrt{\log T}}\right]\right).
\end{align}
Furthermore, by taking $\epsilon=1/T$ and choosing the minimum value of $\Delta^2$ allowed in \autoref{lemma:choice of delta}, \eqref{denguin} yields
\begin{equation*}
\mathcal{V}_T(s)\geq s\sqrt{T \log T} \left(1-\frac{2}{T}\left[1+\frac{s-1}{2s^2}\right]\right).
\end{equation*}
Finally, we need to ensure that $s$ pairs of points $\{z_{i}, z_{i}+w\}$ can be packed in $\mathit{B}_{d}(R)$ such that the distance between any two of the $z_{i}$s is at least $2A$. A sufficient condition \citep{KT1961} is
\begin{equation*}
s \leq \left(\frac{R-2\Delta}{2A\Delta}\right)^d.
\end{equation*}
If $\Delta \leq R/6$ (which is satisfied if $T$ is large enough), the above inequality holds if 
\begin{equation*}
s \leq \left(\frac{R}{3A\Delta}\right)^d %\leq \left(\frac{RT^{1/4}}{6s\sqrt{s(s-1)}\sqrt{\log T}}\right)^d,
%=\left(\frac{\sqrt{2}RT^{\frac{1}{4}}}{3p\sqrt{\log T}}\right)^{d/d+1},  
\end{equation*}
As $A=\sqrt{2}s+1$ and $\Delta^2 < \sqrt{\log T}/\sqrt{T}$, we get the desired result.

\subsection{Proof of \autoref{lemma:convergence of rjmcmc}}

Let $D_{n}$ denote the event that no "within-model move" is ever accepted in the first $n$ moves. Then $D_{1}=D_{1}^{\text{within}}\cup D_{1}^{\text{between}}$, where $D_{1}^{\text{within}}$ stands for the event that a "within-model move" is proposed but rejected in one step and $D_{1}^{\text{between}}$ that a "between-model move" is proposed in one step. Then we have
\begin{align*}
\mathbb{P}\left[D_{1}|(k^{(0)},\bc^{(0)})=(k,\bc)\right]=&\mathbb{P}\left[k^{\prime}\neq k|(k,\bc)\right]+\mathbb{P}\left[k^{\prime}=k, \text{but rejected}|(k,\bc)\right]\\
=&\frac{2}{3}+\frac{1}{3}\left[1-\int_{\mathbb{R}^{dk}}\alpha\left[(k,\bc),(k,\bc^{\prime})\right]\rho_{k}\left(\bc^{\prime},\ttc_{k},\tau_{k}\right)\mathrm{d}\bc^{\prime}\right],
\end{align*}
where 
\begin{align*}
\alpha\left[(k,\bc),(k,\bc^{\prime})\right]&=\min\left\{1,\frac{\hat{\rho}_{t}(\bc^{\prime})\rho_{k}(\bc,\ttc_{k},\tau_{k})}{\hat{\rho}_{t}(\bc)\rho_{k}(\bc^{\prime},\ttc_{k},\tau_{k})}\right\}\\
&=\min\left\{1,h_{t}\left(\bc^{\prime}|(k,\bc)\right)\right\}.
\end{align*}
Under the assumption of $k^{\prime}=k$, we have that $\bc^{\prime},\bc\in\R^{dk}$, therefore the restriction of $\hat{\rho}_{t}$ to $\R^{dk}$ is well defined. Moreover, by the definition of $\pi_{k}$ in \eqref{eq:uniform prior}, the support of the restriction of $\hat{\rho}_{t}$ to $\R^{dk}$ is $\mathbb{R}^{dk}\cap \mathcal{E}=\left(\mathit{B}_{d}(2R)\right)^{k}$. Hence the function $(\bc^{\prime},\bc)\mapsto h_{t}\left(\bc^{\prime}|(k,\bc)\right)$  is strictly positive and continuous on the compact set $\left(\mathit{B}_{d}(2R)\right)^{k} \times \left(\mathit{B}_{d}(2R)\right)^{k}$. As a consequence, the minimum of $h_{t}\left(\bc^{\prime}|(k,\bc)\right)$ on $\left(\mathit{B}_{d}(2R)\right)^{k} \times \left(\mathit{B}_{d}(2R)\right)^{k}$ is achieved and we denote it by $m_{k}$, \emph{i.e.},
\begin{equation*}
m_{k}=\inf_{\bc^{\prime},\bc\in\left(\mathit{B}_{d}(2R)\right)^{k}}h_{t}\left(\bc^{\prime}|(k,\bc)\right)>0.
\end{equation*}
In addition, due to the continuity and positivity of $\rho_{k}$ on $\mathbb{R}^{dk}$, it is clear that for any $k \in \llbracket1,p\rrbracket$
\begin{equation*}
z_{k}=\int_{\left(\mathit{B}_{d}(2R)\right)^{k}}\rho_{k}\left(\bc^{\prime},\ttc_{k},\tau_{k}\right)\mathrm{d}\bc^{\prime} > 0.
\end{equation*}
Therefore, for any $k$,
\begin{align*}
\int_{\mathbb{R}^{dk}}\alpha\left[(k,\bc),(k,\bc^{\prime})\right]\rho_{k}\left(\bc^{\prime},\ttc_{k},\tau_{k}\right)\mathrm{d}\bc^{\prime} &\geq \inf_{k\in\llbracket1,p\rrbracket}\left(m_{{k}}z_{k}\right)\\ &=:m^{\star} >0.
\end{align*}
Hence, uniformly on $k\in \llbracket1,p\rrbracket$ and $\bc \in  \mathbb{R}^{dk}\cap \mathcal{E}$, we have,
\begin{equation*}
\mathbb{P}\left[D_{1}|(k,\bc)\right]\leq \left[\frac{2}{3}+\frac{1}{3}(1-m^{\star})\right] <1.
\end{equation*}
To conclude,
\begin{equation*}
\mathbb{P}\left[D|(k,\bc)\right]=\lim_{n \longrightarrow \infty}\mathbb{P}\left[D_{n}|(k,\bc)\right]\leq\lim_{n \longrightarrow \infty}\left[\frac{2}{3}+\frac{1}{3}(1-m^{\star})\right]^{n}=0.
\end{equation*}

\subsection{Proof of \autoref{thm:theorem 4}}

For any $\bc \in \mathcal{E}$, there exists some $k\in \llbracket1,p\rrbracket$ such that $\bc \in \left(\mathit{B}_{d}(2R)\right)^{k}\subset \mathcal{E}$. For any $k^{\prime} \in \llbracket k-1,k+1\rrbracket$ and for any $A\in \mathcal{B}\left(\mathbb{R}^{dk^{\prime}}\right)$ such that $\hat{\rho}_{t}(A)>0$, the transition kernel $H$ of the chain is given by
\begin{equation}\label{eq:transition kernel of RJMCMC}
H\big(\bc,\bc^{\prime}\in A\big)=\int \mathbbm{1}_{\{v_{1}\in A\}}\alpha\left[\left(k,\bc \right),\left(k^{\prime},v_{1}\right)\right]q(k,k^{\prime})\rho_{k^{\prime}}(v_{1},\ttc_{k^{\prime}},\tau_{k^{\prime}})\mathrm{d}v_{1}+r(\bc)\delta_{\bc}\left(A\right),
\end{equation}
where $\rho_{k^{\prime}}(\cdot,\ttc_{k^{\prime}},\tau_{k^{\prime}})$ is the multivariate Student distribution in \eqref{eq:proposal distribution} and 
\begin{align*}
r\left(\bc\right)=\sum_{k^{\prime}\in \llbracket k-1,k+1\rrbracket}q(k,k^{\prime})\int \left(1-\alpha\left[\left(k,\bc\right),\left(k^{\prime},v_{1}\right)\right]\right)\rho_{k^{\prime}}(v_{1},\ttc_{k^{\prime}},\tau_{k^{\prime}})\mathrm{d}v_{1}
\end{align*}
is the probability of rejection when starting at state $\bc$, and $\delta_{\bc}(\cdot)$ is a Dirac measure in $\bc$. One can easily note that $H(\bc,\bc^{\prime}\in A)$ in \eqref{eq:transition kernel of RJMCMC} is strictly positive, indicating that the chain, when starting from $\bc$, has a positive chance to move. Therefore, for any $A\in \mathcal{B}(\mathcal{C})$ such that $\hat{\rho}_{t}(A)>0$, we can prove with the Chapman-Kolmogorov equation that there exists some $m\in \mathbb{N}^{\ast}$ such that
\begin{equation*}
H^{m}\left(\bc,A\right)>0,
\end{equation*}
where $H^{m}(\bc,A)=\int H^{m-1}(y,A)H(\bc,\mathrm{d}y)$ is the $m$-step transition kernel. In other words, the chain is $\hat{\rho}_{t}$-irreducible.
Finally, a sufficient condition for the chain to be aperiodic is that \autoref{proc:procedure 2} allows transitions such as $\left\{\left(k^{(n+1)},\bc^{(n+1)}\right)=\left(k^{(n)},\bc^{(n)}\right)\right\}$, \emph{i.e.}, 
\begin{equation}\label{eq:aperiodic}
\mathbb{P}\left(\alpha\left[(k^{(n)},\bc^{(n)}),(k^{\prime},\bc^{\prime})\right]<1\right)
=\mathbb{P}\left(\frac{\hat{\rho}_{t}(\bc^{\prime})q(k^{\prime},k^{(n)})\rho_{k^{(n)}}(\bc^{(n)},\ttc_{k^{(n)}},\tau_{k^{(n)}})}{\hat{\rho}_{t}(\bc^{(n)})q(k^{(n)},k^{\prime})\rho_{k^{\prime}}(\bc^{\prime},\ttc_{k^{\prime}},\tau_{k^{\prime}})}<1\right)>0.
\end{equation}
Since for any $\bc^{\prime} \in A\subset \mathcal{B}\left(\mathbb{R}^{dk^{\prime}}\right)\cap \mathcal{E}^{c}$ such that 
%\begin{equation*}
$\mathbb{P}\left(\bc^{\prime}\in A\right)=\int_{A}\rho_{k^{\prime}}(\bc^{\prime},\ttc_{k^{\prime}},\tau_{k^{\prime}})\mathrm{d}\bc^{\prime} >0$,
%\end{equation*}
we have $\hat{\rho}_{t}(\bc^{\prime})=0$, \eqref{eq:aperiodic} holds.
Therefore,
\begin{align*}
\mathbb{P}\left(\frac{\hat{\rho}_{t}(\bc^{\prime})q(k^{\prime},k^{(n)})\rho_{k^{(n)}}(\bc^{(n)},\ttc_{k^{(n)}},\tau_{k^{(n)}})}{\hat{\rho}_{t}(\bc^{(n)})q(k^{(n)},k^{\prime})\rho_{k^{\prime}}(\bc^{\prime},\ttc_{k^{\prime}},\tau_{k^{\prime}})}<1\right)\geq \mathbb{P}\left(\bc^{\prime}\in A\right)>0.
\end{align*}
The chain is therefore aperiodic. Finally, the Harris recurrence of the chain is a consequence of \autoref{lemma:convergence of rjmcmc} \citep[based on][Theorem 20]{RR2006}. As a conclusion, the chain converges to the target distribution $\hat{\rho}_{t}$.

\section*{Acknowledgements}
The authors gratefully acknowledge financial support from iAdvize and ANRT (CIFRE grant 2014-00757).%, and thank an anonymous Referee and the Action Editor for insightful comments.

%\clearpage

\bibliographystyle{plainnat}
\bibliography{biblio}

\begin{thebibliography}{52}
\providecommand{\natexlab}[1]{#1}
\providecommand{\url}[1]{\texttt{#1}}
\expandafter\ifx\csname urlstyle\endcsname\relax
  \providecommand{\doi}[1]{doi: #1}\else
  \providecommand{\doi}{doi: \begingroup \urlstyle{rm}\Url}\fi

\bibitem[Alquier(2006)]{Alq2006}
P.~Alquier.
\newblock \emph{Transductive and Inductive Adaptive Inference for Regression
  and Density Estimation}.
\newblock PhD thesis, Université Paris 6, 2006.

\bibitem[Alquier and Biau(2013)]{AB2013}
P.~Alquier and G.~Biau.
\newblock Sparse single-index model.
\newblock \emph{Journal of Machine Learning Research}, 14:\penalty0 243--280,
  2013.

\bibitem[Alquier and Guedj(2017)]{AG2016}
P.~Alquier and B.~Guedj.
\newblock {An Oracle Inequality for Quasi-Bayesian Non-Negative Matrix
  Factorization}.
\newblock \emph{Mathematical Methods of Statistics}, 2017.

\bibitem[Alquier and Lounici(2011)]{AL2011}
P.~Alquier and K.~Lounici.
\newblock {PAC-Bayesian} theorems for sparse regression estimation with
  exponential weights.
\newblock \emph{Electronic Journal of Statistics}, 5:\penalty0 127--145, 2011.

\bibitem[Audibert(2004)]{Aud2004b}
J.-Y. Audibert.
\newblock \emph{Une approche PAC-bayésienne de la théorie statistique de
  l'apprentissage}.
\newblock PhD thesis, Université Paris 6, 2004.

\bibitem[Audibert(2009)]{Aud2009}
J.-Y. Audibert.
\newblock Fast learning rates in statistical inference through aggregation.
\newblock \emph{The Annals of Statistics}, 37\penalty0 (4):\penalty0
  1591--1646, 2009.

\bibitem[Azoury and Warmuth(2001)]{AW2001}
K.~S. Azoury and M.~K. Warmuth.
\newblock Relative loss bounds for on-line density estimation with the
  exponential family of distributions.
\newblock \emph{Machine Learning}, 43\penalty0 (3):\penalty0 211--246, 2001.

\bibitem[Barbakh and Fyfe(2008)]{BF2008}
W.~Barbakh and C.~Fyfe.
\newblock Online clustering algorithms.
\newblock \emph{International Journal of Neural Systems}, 18\penalty0
  (3):\penalty0 185--194, 2008.

\bibitem[Bartlett et~al.(1998)Bartlett, Linder, and Lugosi]{BLL1998}
P.~L. Bartlett, T.~Linder, and G.~Lugosi.
\newblock The minimax distortion redundancy in empirical quantizer design.
\newblock \emph{IEEE Transactions on Information Theory}, 44\penalty0
  (5):\penalty0 1802--1813, 1998.

\bibitem[Baudry et~al.(2012)Baudry, Maugis, and Michel]{BMM2012}
J.-P. Baudry, C.~Maugis, and B.~Michel.
\newblock Slope heuristics: overview and implementation.
\newblock \emph{Statistics and Computing}, 22\penalty0 (2):\penalty0 455--470,
  2012.

\bibitem[Calinski and Harabasz(1974)]{CH1974}
R.~B. Calinski and J.~Harabasz.
\newblock A dendrite method for cluster analysis.
\newblock \emph{Communications in Statistics}, 3:\penalty0 1--27, 1974.

\bibitem[Catoni(2004)]{Cat2004}
O.~Catoni.
\newblock \emph{{Statistical Learning Theory and Stochastic Optimization}}.
\newblock {\'E}cole d'{\'E}t{\'e} de Probabilit{\'e}s de Saint-Flour 2001.
  Springer, 2004.

\bibitem[Catoni(2007)]{Cat2007}
O.~Catoni.
\newblock \emph{{PAC-Bayesian Supervised Classification: The Thermodynamics of
  Statistical Learning}}, volume~56 of \emph{Lecture notes -- Monograph
  Series}.
\newblock Institute of Mathematical Statistics, 2007.

\bibitem[Cesa-Bianchi(1999)]{CB1999}
N.~Cesa-Bianchi.
\newblock Analysis of two gradient-based algorithms for on-line regression.
\newblock \emph{Journal of Computer and System Sciences}, 59\penalty0
  (3):\penalty0 392--411, 1999.

\bibitem[Cesa-Bianchi and Lugosi(2006)]{CBL2006}
N.~Cesa-Bianchi and G.~Lugosi.
\newblock \emph{Prediction, Learning and Games}.
\newblock Cambridge University Press, New York, 2006.

\bibitem[Cesa-Bianchi et~al.(1996)Cesa-Bianchi, Long, and Warmuth]{CLW1996}
N.~Cesa-Bianchi, P.~M. Long, and M.~K. Warmuth.
\newblock Worst-case quadratic loss bounds for prediction using linear
  functions and gradient descent.
\newblock \emph{IEEE Transactions on Neural Networks}, 7\penalty0 (3):\penalty0
  604--619, 1996.

\bibitem[Cesa-Bianchi et~al.(1997)Cesa-Bianchi, Helmbold, Freund, Haussler, and
  Warmuth]{CFH1997}
N.~Cesa-Bianchi, D.~Helmbold, N.~Freund, Y.~Haussler, and M.~K. Warmuth.
\newblock How to use expert advice.
\newblock \emph{Journal of the ACM}, 44\penalty0 (3):\penalty0 427--485, 1997.

\bibitem[Cesa-Bianchi et~al.(2007)Cesa-Bianchi, Mansour, and Stoltz]{CMS2007}
N.~Cesa-Bianchi, Y.~Mansour, and G.~Stoltz.
\newblock Improved second-order bounds for prediction with expert advice.
\newblock \emph{Machine Learning}, 66\penalty0 (2):\penalty0 321--352, 2007.
\newblock ISSN 1573-0565.
\newblock \doi{10.1007/s10994-006-5001-7}.
\newblock URL \url{https://doi.org/10.1007/s10994-006-5001-7}.

\bibitem[Choromanska and Monteleoni(2012)]{CM2012}
A.~Choromanska and C.~Monteleoni.
\newblock Online clustering with experts.
\newblock In \emph{Proceedings of the 15th International Conference on
  Artificial Intelligence and Statistics (AISTATS)}, pages 227--235, 2012.

\bibitem[Csisz\'ar(1975)]{C75}
I.~Csisz\'ar.
\newblock I-divergence geometry of probability distributions and minimization
  problems.
\newblock \emph{Annals of Probability}, 3:\penalty0 146--158, 1975.

\bibitem[Dalalyan and Tsybakov(2007)]{DT2007}
A.~S. Dalalyan and A.~B. Tsybakov.
\newblock Aggregation by exponential weighting and sharp oracle inequalities.
\newblock In \emph{Learning theory (COLT 2007), Lecture Notes in Computer
  Science}, pages 97--111, 2007.

\bibitem[Dalalyan and Tsybakov(2008)]{DT2008}
A.~S. Dalalyan and A.~B. Tsybakov.
\newblock Aggregation by exponential weighting, sharp {PAC-Bayesian} bounds and
  sparsity.
\newblock \emph{Machine Learning}, 72:\penalty0 39--61, 2008.

\bibitem[Dalalyan and Tsybakov(2012)]{dalalyan2012sparse}
A.~S. Dalalyan and A.~B. Tsybakov.
\newblock {Sparse regression learning by aggregation and Langevin Monte-Carlo}.
\newblock \emph{Journal of Computer and System Sciences}, 78\penalty0
  (5):\penalty0 1423--1443, 2012.

\bibitem[Dellaportas et~al.(2002)Dellaportas, Forster, and Ntzoufras]{DFN2002}
P.~Dellaportas, J.~J. Forster, and I.~Ntzoufras.
\newblock On {Bayesian} model and variable selection using {MCMC}.
\newblock \emph{Statistics and Computing}, 12\penalty0 (1):\penalty0 27--36,
  2002.

\bibitem[Fischer(2011)]{Fis2011}
A.~Fischer.
\newblock On the number of groups in clustering.
\newblock \emph{Statistics and Probability Letters}, 81:\penalty0 1771--1781,
  2011.

\bibitem[Gerchinovitz(2011)]{Ger2011}
S.~Gerchinovitz.
\newblock \emph{Prédiction de suites individuelles et cadre statistique
  classique : étude de quelques liens autour de la régression parcimonieuse
  et des techniques d'agrégation}.
\newblock PhD thesis, Université Paris-Sud, 2011.

\bibitem[Gordon(1999)]{Gor1999}
A.~D. Gordon.
\newblock \emph{Classification}, volume~82 of \emph{Monographs on Statistics
  and Applied Probability}.
\newblock Chapman Hall/CRC, Boca Raton, 1999.

\bibitem[Green(1995)]{Gre1995}
P.~J. Green.
\newblock {Reversible Jump Markov Chain Monte Carlo computation and Bayesian
  model determination}.
\newblock \emph{Biometrika}, 82\penalty0 (4):\penalty0 711--732, 1995.

\bibitem[Guedj and Alquier(2013)]{GA2013}
B.~Guedj and P.~Alquier.
\newblock {PAC-Bayesian} estimation and prediction in sparse additive models.
\newblock \emph{Electronic Journal of Statistics}, 7:\penalty0 264--291, 2013.

\bibitem[Guedj and Robbiano(2017)]{GR2015}
B.~Guedj and S.~Robbiano.
\newblock {PAC-Bayesian high dimensional bipartite ranking}.
\newblock \emph{Journal of Statistical Planning and Inference}, 2017.

\bibitem[Guha et~al.(2003)Guha, Meyerson, Mishra, Motwani, and
  O'Callaghan]{GMM2003}
S.~Guha, A.~Meyerson, N.~Mishra, R.~Motwani, and L.~O'Callaghan.
\newblock Clustering data streams: theory and practice.
\newblock \emph{IEEE Transactions on Knowledge and Data Engineering},
  15\penalty0 (3):\penalty0 511--528, 2003.

\bibitem[Hartigan(1975)]{Har1975}
J.~A. Hartigan.
\newblock \emph{Clustering Algorithms}.
\newblock Wiley Series in Probability and Mathematical Statistics. John Wiley
  and Sons, New York, 1975.

\bibitem[Kaufman and Rousseeuw(1990)]{KR1990}
L.~Kaufman and P.~Rousseeuw.
\newblock \emph{Finding Groups in Data: An Introduction to Cluster Analysis}.
\newblock Wiley Series in Probability and Mathematical Statistics.
  Wiley-Interscience, Hoboken, 1990.

\bibitem[Kivinen and Warmuth(1997)]{KW1997}
J.~Kivinen and M.~K. Warmuth.
\newblock Exponentiated gradient versus gradient descent for linear predictors.
\newblock \emph{Information and Computation}, 132\penalty0 (1):\penalty0 1--63,
  1997.

\bibitem[Kivinen and Warmuth(1999)]{KW1999}
J.~Kivinen and M.~K. Warmuth.
\newblock Averaging expert predictions.
\newblock In \emph{Computational Learning Theory: 4th European Conference
  (EuroCOLT ’99)}, pages 153--167. Springer, 1999.

\bibitem[Kolmogorov and Tikhomirov(1961)]{KT1961}
A.~N. Kolmogorov and V.~M. Tikhomirov.
\newblock $\epsilon$-entropy and $\epsilon$-capacity of sets in function
  spaces.
\newblock \emph{American Mathematical Society Translations}, 17:\penalty0
  277--364, 1961.

\bibitem[Krzanowski and Lai(1988)]{KL1985}
W.~J. Krzanowski and Y.~T. Lai.
\newblock A criterion for determination the number of clusters in a data set.
\newblock \emph{Biometrics}, 44:\penalty0 23--34, 1988.

\bibitem[Li(2016)]{R-PACBO}
L.~Li.
\newblock \emph{{PACBO}: PAC-Bayesian Online Clustering}, 2016.
\newblock URL \url{https://CRAN.R-project.org/package=PACBO}.
\newblock R package version 0.1.0.

\bibitem[Liberty et~al.(2016)Liberty, Sriharsha, and Sviridenko]{LSS2015}
E.~Liberty, R.~Sriharsha, and M.~Sviridenko.
\newblock An algorithm for online $k$-means clustering.
\newblock In \emph{Proceedings of the Eighteenth Workshop on Algorithm
  Engineering and Experiments (ALENEX)}, pages 81--89. SIAM, 2016.

\bibitem[Littlestone and Warmuth(1994)]{LW1994}
N.~Littlestone and M.~K. Warmuth.
\newblock The weighted majority algorithm.
\newblock \emph{Information and Computation}, 108\penalty0 (2):\penalty0
  212--216, 1994.

\bibitem[McAllester(1999{\natexlab{a}})]{McA1998}
D.~A. McAllester.
\newblock Some {PAC-Bayesian} theorems.
\newblock \emph{Machine Learning}, 37\penalty0 (3):\penalty0 355--363,
  1999{\natexlab{a}}.

\bibitem[McAllester(1999{\natexlab{b}})]{McA1999}
D.~A. McAllester.
\newblock {PAC-Bayesian} model averaging.
\newblock In \emph{Proceedings of the 12th annual conference on Computational
  Learning Theory}, pages 164--170. ACM, 1999{\natexlab{b}}.

\bibitem[Milligan and Cooper(1985)]{MC1985}
G.~W. Milligan and M.~C. Cooper.
\newblock An examination of procedures for determining the number of clusters
  in a data set.
\newblock \emph{Psychometrika}, 50:\penalty0 159--179, 1985.

\bibitem[Petralias and Dellaportas(2013)]{PD2012}
A.~Petralias and P.~Dellaportas.
\newblock {An MCMC model} search algorithm for regression problems.
\newblock \emph{Journal of Statistical Computation and Simulation}, 83\penalty0
  (9):\penalty0 1722--1740, 2013.

\bibitem[Robert and Casella(2004)]{RG2004}
C.~P. Robert and G.~Casella.
\newblock \emph{Monte Carlo Statistical Methods}.
\newblock Springer, New York, 2004.

\bibitem[Roberts and Rosenthal(2006)]{RR2006}
G.~O. Roberts and J.~S. Rosenthal.
\newblock {Harris Recurrence of Metropolis-Within-Gibbs and Trans-Dimensional
  Markov Chains}.
\newblock \emph{Annals of Applied Probability}, 16\penalty0 (4):\penalty0
  2123--2139, 2006.

\bibitem[Seeger(2002)]{See2002}
M.~Seeger.
\newblock {PAC-Bayesian} generalization bounds for gaussian processes.
\newblock \emph{Journal of Machine Learning Research}, 3:\penalty0 233--269,
  2002.

\bibitem[Seeger(2003)]{See2003}
M.~Seeger.
\newblock \emph{Bayesian Gaussian Process Models: {PAC-Bayesian} Generalisation
  Error Bounds and Sparse Approximations}.
\newblock PhD thesis, University of Edinburgh, 2003.

\bibitem[Shawe-Taylor and Williamson(1997)]{STW1997}
J.~Shawe-Taylor and R.~C. Williamson.
\newblock {A PAC analysis of a Bayes estimator}.
\newblock In \emph{Proceedings of the 10th annual conference on Computational
  Learning Theory}, pages 2--9. ACM, 1997.
\newblock \doi{10.1145/267460.267466}.

\bibitem[Tibshirani et~al.(2001)Tibshirani, Walther, and Hastie]{TWH2001}
R.~Tibshirani, G.~Walther, and T.~Hastie.
\newblock Estimating the number of clusters in a dataset via the gap statistic.
\newblock \emph{Journal of the Royal Statistical Society}, 63:\penalty0 411--
  423, 2001.

\bibitem[Vovk(2001)]{Vov2001}
V.~Vovk.
\newblock Competitive on-line statistics.
\newblock \emph{International Statistical Review}, 69\penalty0 (2):\penalty0
  213--248, 2001.

\bibitem[Wintenberger(2017)]{Win2017}
O.~Wintenberger.
\newblock {Optimal learning with Bernstein online aggregation}.
\newblock \emph{Machine Learning}, 106\penalty0 (1):\penalty0 119--141, 2017.

\end{thebibliography}

% \clearpage

\appendix
\section{Extension to a different prior}\label{app}

For the sake of completion, this appendix presents additional regret bounds for a different heavy-tailed prior. Doing so, we stress that the quasi-Bayesian approach is flexible in the sense that it allows for regret bounds for a large variety of priors.
\medskip

Let us consider $\pi_{k}$ as a product of $k$ independent truncated multivariate Student distributions with 3 degrees of freedom in $\R^d$, namely, for any $\bc \in \R^{dk} \subset \mathcal{C}$,
\begin{equation}\label{eq:t distribution on Rdk}
\mathrm{d}\pi_{k}(\bc, \tau_{0},2R)=\prod_{j=1}^{k}\left\{C_{2R,\tau_{0}}^{-1}\left(1+\frac{|c_{j}|^2_{2}}{6\tau_{0}^2}\right)^{-\frac{3+d}{2}}\mathbbm{1}_{\{|c_{j}|_{2}\leq 2R\}}\right\}\mathrm{d}\bc,
\end{equation}
where $\tau_{0}>0$ and $R>0$ are respectively the scale and truncation parameters, and $C_{2R,\tau_{0}}$ is the normalizing constant accounting for the truncation.
When $R=+\infty$, $\pi_{k}(\bc,\tau_{0},2R)$ amounts to a distribution without truncation. In the following, we shorten $\pi_{k}(\bc,\tau_{0},2R)$ to $\pi_{k}$ whenever no confusion is possible.
\medskip

Denote by $\nu$ the multivariate Student distribution in $\R^{d}$, with mean vector $0\in\R^{d}$, scale parameter 1, and 3 degrees of freedom. Fix $k \in \llbracket 1,p\rrbracket$, $R>0$ and $\ttc \in \mathcal{C}(k,R)$, and recall that 
$\Xi (k,R)$ denotes the hypercube in $\R^{k}$ defined by
\begin{equation*}
\Xi (k,R):=\left\{\xi =(\xi_{j})_{j=1,\dots,k}\in\R^{k}\colon  0<\xi_{j}\leq R, \forall j\right\}.
\end{equation*}
For any $k \in \llbracket 1,p\rrbracket$, $\bc\in \R^{dk} \subset \mathcal{C}$, $\ttc \in \mathcal{C}(k,R)$, $\xi \in \Xi(k,R)$, $0<\tau^{2}\leq \sqrt{3}R^{2}/(6\sqrt{d})$ and $R>0$, we define the probability distribution $\rho_{k}$ on $\R^{dk}$ by
\begin{equation}\label{eq:particular t distribution}
\rho_{k}(\bc,\ttc,\tau,\xi)=\prod_{j=1}^{k}\left\{C_{\xi_{j},\tau}^{-1}\left(1+\frac{|c_{j}-\ttc_{j}|^2_{2}}{6\tau^2}\right)^{-\frac{3+d}{2}}\mathbbm{1}_{\{|c_{j}-\ttc_{j}|_{2}\leq \xi_{j}\}}\right\},
\end{equation}
where $C_{\xi_{j},\tau}$ are normalizing constants defined as $C_{\xi_{j},\tau}=\mathbb{P}\left(|\nu|_{2}\leq \xi_{j}/\sqrt{2}\tau\right)/A_{d,\tau}$, where $A_{d,\tau}$ is the constant in the density of $\nu$. Moreover, when $(\xi_{j})_{j=1,\dots,k}=+\infty$, we let  $\rho_{k}(\bc,\ttc,\tau,\xi)$ denote the multivariate Student distribution without truncation.
In the sequel, we will shorten $\rho_{k}(\bc,\ttc,\tau,\xi)$ to $\rho_{k}$ whenever no confusion is possible.
\begin{lem}\label{lem:lemma 1}
Assume that $q$ and $\pi_{k}$ in \eqref{eq:prior distribution} are defined respectively as in \eqref{eq:prior on finite set} and \eqref{eq:t distribution on Rdk}, and that $\rho_{k}$ is defined as \eqref{eq:particular t distribution} for each $k \in \llbracket 1,p\rrbracket$. For the probability distribution $\rho(\bc,\ttc,\tau,\xi)=\mathbbm{1}_{\{\bc\in \R^{dk}\}}\rho_{k}(\bc,\ttc,\tau,\xi)$ defined on $\mathcal{C}$, if $R\geq\max_{t=1,\dots,T}|x_{t}|_{2}$, then
\begin{align*}
\K(\rho,\pi)&\leq \sum_{j=1}^{k}\left[\frac{3+d}{2} \log \left(1+\frac{\xi_{j}^{2}}{6\tau^{2}}\right)-\frac{d}{2} \log \xi_{j}^{2}\right]-k \log c_{d}\\
&+(3+d)k \log\left(1+\frac{\tau}{\tau_{0}}+\frac{\sum_{j=1}^{k}|\ttc_{j}|_{2}}{\sqrt{6}k\tau_{0}}\right)+kd \log \tau_{0}+ \log p+\eta(k-1).\\
\end{align*}
\end{lem}

\begin{proof}
By the definition of the Kullback-Leibler divergence, we have
\begin{align}\label{eq:kullback student}
\K(\rho,\pi)=\K(\rho_{k},\pi_{k})+\log\frac{1}{q(k)}=:A+B,
\end{align}
where
\begin{align}\label{eq:inter kullback student}
A&=\int_{\R^{dk}}\log\left[\prod_{j=1}^{k}\frac{C_{2R,\tau_{0}}}{C_{\xi_{j},\tau}}\left(\frac{\tau_{0}^{2}}{\tau^{2}}\frac{6\tau^{2}+|c_{j}-\ttc_{j}|_{2}^{2}}{6\tau_{0}^{2}+|c_{j}|_{2}^{2}}\right)^{-\frac{3+d}{2}}\right]\rho_{k}(\bc)\mathrm{d}\bc\notag\\
&=\sum_{j=1}^{k}\log\frac{C_{2R,\tau_{0}}}{C_{\xi_{j},\tau}}+\frac{3+d}{2}\int_{\R^{dk}}\sum_{j=1}^{k}\log\left(\frac{\tau^{2}}{\tau_{0}^{2}}\frac{6\tau_{0}^{2}+|c_{j}|_{2}^{2}}{6\tau^{2}+|c_{j}-\ttc_{j}|_{2}^{2}}\right)\rho_{k}(\bc)\mathrm{d}\bc\notag\\
&=\sum_{j=1}^{k}\log\frac{\mathbb{P}\left(|\nu|_{2}\leq \frac{2R}{\sqrt{2}\tau_{0}}\right)}{\mathbb{P}\left(|\nu)|_{2}\leq\frac{\xi_{j}}{\sqrt{2}\tau}\right)}+kd\log\frac{\tau_{0}}{\tau}
+\frac{3+d}{2}\int_{\R^{dk}}\sum_{j=1}^{k}\log\left(\frac{\tau^{2}}{\tau_{0}^{2}}\frac{6\tau_{0}^{2}+|c_{j}|_{2}^{2}}{6\tau^{2}+|c_{j}-\ttc_{j}|_{2}^{2}}\right)\rho_{k}(\bc)\mathrm{d}\bc \notag\\ &=:A_{1}+A_{2}+A_{3}.
\end{align}
By the definition of the multivariate Student distribution $\nu$,
\begin{align*}
\mathbb{P}\left(|\nu|_{2}\leq \frac{\xi_{j}}{\sqrt{2}\tau}\right)&=\int_{|\nu|_{2}\leq \frac{\xi_{j}}{\sqrt{2}\tau}}\frac{\Gamma(\frac{3+d}{2})}{\Gamma(\frac{3}{2})(3\pi)^{\frac{d}{2}}}\left(1+\frac{|\nu|_{2}^{2}}{3}\right)^{-\frac{3+d}{2}}\mathrm{d}\nu\\
&\geq \left(1+\frac{\xi_{j}^{2}}{6\tau^{2}}\right)^{-\frac{3+d}{2}}\frac{\Gamma(\frac{3+d}{2})}{\Gamma(\frac{3}{2})(3\pi)^{\frac{d}{2}}}\int_{|\nu|_{2}\leq \frac{\xi_{j}}{\sqrt{2}\tau}}\mathrm{d}\nu\\
&=c_{d}\tau^{-d}\left(1+\frac{\xi_{j}^{2}}{6\tau^{2}}\right)^{-\frac{3+d}{2}}\xi_{j}^{d},
\end{align*}
where $\Gamma(\cdot)$ is the Gamma function and
$
c_{d}=\frac{\Gamma\left(\frac{3+d}{2}\right)}{\Gamma\left(\frac{3}{2}\right)\Gamma\left(\frac{d}{2}+1\right)6^{\frac{d}{2}}}.
$
Hence, the term $A_{1}$ in \eqref{eq:inter kullback student} verifies
\begin{align}\label{ieq:first term student}
A_{1}&=k\log\mathbb{P}\left(|\nu|_{2}\leq \frac{2R}{\sqrt{2}\tau_{0}}\right)-\sum_{j=1}^{k}\log \mathbb{P}\left(|\nu)|_{2}\leq\frac{\xi_{j}}{\sqrt{2}\tau}\right)\notag\\
&\leq-\sum_{j=1}^{k}\log \mathbb{P}\left(|\nu|_{2}\leq \frac{\xi_{j}}{\sqrt{2}\tau}\right)\notag\\
&\leq\sum_{j=1}^{k}\left[\frac{3+d}{2}\log\left(1+\frac{\xi_{j}^{2}}{6\tau^{2}}\right)-\frac{d}{2}\log \xi_{j}^{2}\right]+kd\log \tau-k\log c_{d}.
\end{align}
In addition, we have
\begin{align*}
\frac{6\tau^{2}_{0}+|c_{j}|_{2}^{2}}{6\tau^{2}+|c_{j}-\ttc_{j}|_{2}^{2}}
&\leq 1+\frac{2|\ttc_{j}|_{2}}{2\sqrt{6}\tau}\frac{2\sqrt{6}\tau|c_{j}-\ttc_{j}|_{2}}{6\tau^{2}+|c_{j}-\ttc_{j}|_{2}^{2}}+\frac{|\ttc_{j}|_{2}^2}{6\tau^{2}+|c_{j}-\ttc_{j}|_{2}^{2}}+\frac{\tau_{0}^{2}}{\tau^{2}}\notag\\
&=1+\frac{|\ttc_{j}|_{2}}{\sqrt{6}\tau}+\frac{|\ttc_{j}|_{2}^2}{6\tau^{2}}+\frac{\tau_{0}^{2}}{\tau^{2}}\leq \left(1+\frac{|\ttc_{j}|_{2}}{\sqrt{6}\tau}+\frac{\tau_{0}}{\tau}\right)^{2},
\end{align*}
where we used the Cauchy–Schwarz inequality. Due to the above inequality, the term $A_{3}$ in \eqref{eq:inter kullback student} satisfies
\begin{align}\label{ieq:second term student}
A_{3}&\leq (3+d)\int\sum_{j=1}^{k}\log\left(1+\frac{\tau}{\tau_{0}}+\frac{|\ttc_{j}|_{2}}{\sqrt{6}\tau_{0}}\right)\rho_{k}(\bc)\mathrm{d}\bc\notag\\
&\leq (3+d)k\int\log\left(1+\frac{\tau}{\tau_{0}}+\frac{\sum_{j=1}^{k}|\ttc_{j}|_{2}}{\sqrt{6}k\tau_{0}}\right)\rho_{k}(\bc)\mathrm{d}\bc\notag\\
&=(3+d)k\log\left(1+\frac{\tau}{\tau_{0}}+\frac{\sum_{j=1}^{k}|\ttc_{j}|_{2}}{\sqrt{6}k\tau_{0}}\right).
\end{align}
Combining \eqref{eq:kullback student}, \eqref{eq:inter kullback student}, \eqref{ieq:first term student}, \eqref{ieq:second term student} with \eqref{ieq:log discrete probability} completes the proof.
\end{proof}

\begin{coro}\label{cor:corollary 1 student}
For any sequence $(x_{t})_{1:T} \in \R^{dT}$, for any $\lambda>0$, if $q$ and $\pi_{k}$ in \eqref{eq:prior distribution} are taken respectively as in \eqref{eq:prior on finite set} and \eqref{eq:t distribution on Rdk} with parameter $\eta\geq 0$, $\tau_{0}>0$ and $R\geq \max_{t=1,\dots,T}|x_{t}|_{2}$, \autoref{proc:procedure 1} satisfies, for any $0<\tau^{2}\leq (\sqrt{3}R^{2})/(6\sqrt{d})$,
\begin{align*}
&\sum_{t=1}^T\mathbb{E}_{(\hat{\rho}_{1},\hat{\rho}_{2},\dots,\hat{\rho}_{t})}\ell(\hat{\bc}_{t},x_{t})\leq\inf_{k\in \llbracket 1,p\rrbracket}\inf_{\bc\in\mathcal{C}(k,R)}\left\{\sum_{t=1}^{T}\ell(\bc,x_{t})+\frac{kd}{\lambda}\log \frac{\tau_{0}}{ c_{d}\tau}+\frac{\eta}{\lambda}k \right.\\
&\left.+\frac{(3+d)k}{\lambda}\log\left(1+\frac{\tau}{\tau_{0}}+\frac{\sum_{j=1}^{k}|c_{j}|_{2}}{\sqrt{6}k\tau_{0}}\right)+\frac{1}{\lambda}\sqrt{kd(12\tau^2T\lambda+3k)}\right\}+\frac{\lambda T}{2}C_{1}^2+\frac{\log p}{\lambda},
\end{align*}
where $C_{1}=(2R+\max_{t=1,\dots,T}|x_{t}|_{2})^2$ and  $c_{d}=\left(\frac{\Gamma(\frac{3+d}{2})}{\Gamma(\frac{3}{2})\Gamma(\frac{d}{2}+1)}\right)^{1/d}$.
\end{coro}

\begin{proof}
By \autoref{thm:theorem 1},
\begin{align}\label{prf:corollary 1 student}
\expcum\leq&\inf_{k\in \llbracket 1,p\rrbracket}\inf_{\substack{\rho\in \PP_\pi(\mathcal{C})\\\rho=\rho_{k}\mathbbm{1}_{\{\bc\in\R^{dk}\}}}}\left\{\mathbb{E}_{\bc\sim \rho}\sum_{t=1}^{T}[\ell(\bc,x_{t})]+\frac{\K(\rho,\pi)}{\lambda}\right. \notag\\
&\left.+\frac{\lambda}{2}\mathbb{E}_{(\hat{\rho}_{1},\dots,\hat{\rho}_{T})}\mathbb{E}_{\bc\sim \rho}\sum_{t=1}^{T}[\ell(\bc,x_{t})-\ell(\hat{\bc}_{t},x_{t})]^2\right\}\notag\\
\end{align}
As in \eqref{prf:expected part 1}, the first term on the right-hand side of \eqref{prf:corollary 1 student} may be upper bounded.
\begin{align}\label{prf:expected part 1 student}
\sum_{t=1}^{T}\mathbb{E}_{\bc\sim \rho}[\ell(\bc,x_{t})]\leq \sum_{t=1}^{T}\ell(m,x_{t})+T\max_{j=1,\dots,k}\xi_{j}^{2}.
\end{align}
For the second term in the right-hand side of \eqref{prf:corollary 1 student}, by \autoref{lem:lemma 1},
\begin{align}\label{prf:second term student}
\frac{\K(\rho,\pi)}{\lambda}
&\leq \frac{(3+d)k}{\lambda}\log\left(1+\frac{\tau}{\tau_{0}}+\frac{\sum_{j=1}^{k}|\ttc_{j}|_{2}}{\sqrt{6}k\tau_{0}}\right)+\frac{1}{\lambda}\sum_{j=1}^{k}\left[\frac{3+d}{2}\log\left(1+\frac{\xi_{j}^{2}}{6\tau^{2}}\right)-\frac{d}{2}\log \xi_{j}^{2}\right]\notag\\
&+\frac{kd}{\lambda}\log \tau_{0}-\frac{k}{\lambda}\log c_{d}+\frac{\eta}{\lambda}(k-1)+\frac{\log p}{\lambda}.
\end{align}
Likewise to \eqref{prf:third term}, the third term on the right-hand side of \eqref{prf:corollary 1 student} is upper bounded by
\begin{align}\label{prf:third term student}
\frac{\lambda}{2}\mathbb{E}_{(\hat{\rho}_{1},\dots,\hat{\rho}_{T})}\mathbb{E}_{\bc\sim \rho_{k}}\sum_{t=1}^{T}[\ell(\bc,x_{t})-\ell(\hat{\bc}_{t},x_{t})]^2
\leq \frac{\lambda T}{2}C_{1}^{2}.
\end{align}
Combining inequalities \eqref{prf:expected part 1 student}, \eqref{prf:second term student} and \eqref{prf:third term student} yields for $\xi\in \Xi(k,R)$ and $0<\tau^{2}\leq \sqrt{3}R^{2}/(6\sqrt{d})$ that
\begin{align*}
&\expcum
\leq\inf_{k\in \llbracket 1,p\rrbracket}\inf_{\ttc\in\mathcal{C}(k,R)}\left\{\sum_{t=1}^{T}\ell(\ttc,x_{t})+\xi_{j}^{2}+\frac{(3+d)k}{\lambda}\log\left(1+\frac{\tau}{\tau_{0}}+\frac{\sum_{j=1}^{k}|\ttc_{j}|_{2}}{\sqrt{6}k\tau_{0}}\right) \right. \notag\\
&\left. +T\max_{j=1,\dots,k}\xi_j^2 +\frac{3+d}{2\lambda}\sum_{j=1}^{k}\log \left(1+\frac{\xi_{j}^2}{6\tau^{2}}\right)-\frac{d}{2\lambda}\sum_{j=1}^{k}\log \xi_{j}^{2}+\frac{kd}{\lambda}\log \tau_{0}-\frac{k}{\lambda}\log c_{d}+(k-1)\right\}\notag\\
&+\frac{\lambda T}{2}C_{1}^{2}+\frac{\log p}{\lambda}.
\end{align*}
Let  $\hat{\xi}_{j}=\xi_{j}^{2}/6\tau^{2}$ for any $j=1,\dots,k$, then $0 < \hat{\xi}_{j}\leq R^2/6\tau^2$ since $\xi=(\xi_{j})_{j=1,\dots,k}\in\Xi (k,R)$. This yields
\begin{align}\label{ieq:inequality of xi}
&T\max_{j=1,\dots,k}\xi_{j}^{2}+\frac{3+d}{2\lambda}\sum_{j=1}^{k}\log \left(1+\frac{\xi_{j}^2}{6\tau^2}\right)-\frac{d}{2\lambda}\sum_{j=1}^{k}\log \xi_{j}^{2}\notag\\
&=6\tau^2T\max_{j=1,\dots,k}\hat{\xi}_{j}+\frac{3}{2\lambda}\sum_{j=1}^{k}\log \left(1+\hat{\xi}_{j}\right)+\frac{d}{2\lambda}\sum_{j=1}^{k}\log \left(1+\frac{1}{\hat{\xi}_{j}}\right)-\frac{kd}{2\lambda}\log(6\tau^{2})\notag\\
&\leq 6\tau^2T\max_{j=1,\dots,k}\hat{\xi}_{j}+\frac{3}{2\lambda}\sum_{j}^{k}\hat{\xi}_{j}+\frac{d}{2\lambda}\sum_{j=1}^{k}\frac{1}{\hat{\xi}_{j}}-\frac{kd}{2\lambda}\log(6\tau^{2})\notag\\
&\leq \left(6\tau^2T+\frac{3k}{2\lambda}\right)\max_{j=1,\dots,k}\hat{\xi}_{j}+\frac{d}{2\lambda}\sum_{j=1}^{k}\frac{1}{\hat{\xi}_{j}}-\frac{kd}{2\lambda}\log(6\tau^{2}).
\end{align}
The minimum of the right-hand side of \eqref{ieq:inequality of xi} is reached for
\begin{equation*}
\hat{\xi}_{1}=\dots=\hat{\xi}_{k}=\sqrt{\frac{kd}{12\tau^{2}T\lambda+3k}} \leq \frac{R^2}{6\tau^2}, \quad\textrm{if }0<\tau^{2}\leq \frac{\sqrt{3}R^{2}}{6\sqrt{d}}.
\end{equation*}
Therefore for a fixed $k$, $\ttc\in\mathcal{C}(k,R)$ and $0<\tau^{2}\leq \frac{\sqrt{3}R^{2}}{6\sqrt{d}}$,
\begin{multline*}
\inf_{{\xi}\in\Xi (k,R)}\left\{T\max_{j=1,\dots,k}\xi_{j}^{2}+\frac{3+d}{2\lambda}\sum_{j=1}^{k}\log \left(1+\frac{\xi_{j}^2}{6\tau^2}\right)-\frac{d}{2\lambda}\sum_{j=1}^{k}\log \xi_{j}^{2}\right\}\leq \frac{1}{\lambda}\sqrt{kd(12\tau^2T\lambda+3k)} \\ -\frac{kd}{2\lambda}\log 6\tau^2.
\end{multline*}
Hence
\begin{align*}
&\expcum
\leq\inf_{k\in \llbracket 1,p\rrbracket}\inf_{\ttc\in\mathcal{C}(k,R)}\left\{\sum_{t=1}^{T}\ell(\ttc,x_{t})+\frac{(3+d)k}{\lambda}\log\left(1+\frac{\tau}{\tau_{0}}+\frac{\sum_{j=1}^{k}|\ttc_{j}|_{2}}{\sqrt{6}k\tau_{0}}\right)\right.\\
&\left.+\frac{1}{\lambda}\sqrt{kd(12\tau^2T\lambda+3k)}+\frac{kd}{\lambda}\log \frac{\tau_{0}}{\sqrt{6}\tau c_{d}^{1/d}}+\frac{\eta}{\lambda}(k-1)\right\}+\frac{\lambda T}{2}C_{1}^2+\frac{\log p}{\lambda}.
\end{align*}
which concludes the proof.
\end{proof}

Tuning parameters $\lambda$, $\tau$ and $\eta$ can be chosen to obtain a sublinear regret bound for the cumulative loss of \autoref{proc:procedure 1}.

\begin{coro}\label{cor:corollary 2 student}
For any sequence $(x_{t})_{1:T} \in \R^{dT}$, under the assumptions of \autoref{cor:corollary 1 student}, if $T\geq 12d\tau_{0}^{4}/c_{d}^{2}R^{4}$,  $\lambda=\sqrt{\log T}/\sqrt{T}$,  $\tau^{2}=\tau_{0}^{2}T^{-1/2}(c_{d})^{-2}$ and $\eta\geq 0$, \autoref{proc:procedure 1} satisfies
\begin{align*}
&\expcum\leq\inf_{k\in \llbracket 1,p\rrbracket}\inf_{\bc\in\mathcal{C}(k,R)}\left\{\sum_{t=1}^{T}\ell(\bc,x_{t})+(3+d)k\sqrt{T}\log\left(1+\frac{1}{c_{d}T^{\frac{1}{4}}}+\frac{\sum_{j=1}^{k}|c_{j}|_{2}}{\sqrt{6}k\tau_{0}}\right)\right.\notag\\
&\left.+\frac{kd}{4}\sqrt{T\log T}+\left(\sqrt{3k^2d+12\tau_{0}^{2}(c_{d})^{-2}}+\eta k\right)\sqrt{T}\right\}+\left(\log p+\frac{C_{1}^{2}}{2}\right)\sqrt{T},
\end{align*}
where $C_{1}=(2R+\max_{t=1,\dots,T}|x_{t}|_{2})^2$ and  $c_{d}=\left(\frac{\Gamma(\frac{3+d}{2})}{\Gamma(\frac{3}{2})\Gamma(\frac{d}{2}+1)}\right)^{1/d}$.
\end{coro}
In the adaptive setting (\autoref{proc:alg1bis}), applying \autoref{thm:theorem 2} to the specific $q$ and $\pi_{k}$ in \eqref{eq:prior on finite set} and \eqref{eq:t distribution on Rdk} leads to the following result.

\begin{coro}\label{cor:corollary 3 student}
For any deterministic sequence $(x_{t})_{1:T} \in \R^{dT}$, under the assumptions of \autoref{cor:corollary 1 student}, set $T\geq 12d\tau_{0}^{4}/c_{d}^{2}R^{4}$, $\eta\geq 0$, $R\geq \max_{t=1,\dots,T}|x_{t}|_{2}$ and $\lambda_{t}=\sqrt{\log t}/\sqrt{t}$ for any $t\in\llbracket 1,T\rrbracket$ and $\lambda_0=1$. Then \autoref{proc:alg1bis} satisfies
\begin{multline*}
\expcum
\leq\inf_{k\in \llbracket 1,p\rrbracket}\inf_{\bc\in\mathcal{C}(k,R)}\left\{\sum_{t=1}^{T}\ell(\bc,x_{t})+(3+d)k\sqrt{T} \log\left(1+\frac{1}{c_{d}T^{\frac{1}{4}}}+\frac{\sum_{j=1}^{k}|c_{j}|_{2}}{\sqrt{6}k\tau_{0}}\right) \right. \notag\\ \left.
+\frac{kd}{4}\sqrt{T \log T} +\left(\sqrt{3k^2d+12\tau_{0}^{2}(c_{d})^{-2}}+\eta k\right)\sqrt{T}\right\}+\left( \log p+C_{1}^{2}\right)\sqrt{T},
\end{multline*}
where $C_{1}=(2R+\max_{t=1,\dots,T}|x_{t}|_{2})^2$ and  $c_{d}=\left(\frac{\Gamma(\frac{3+d}{2})}{\Gamma(\frac{3}{2})\Gamma(\frac{d}{2}+1)}\right)^{1/d}$.
\end{coro}

\begin{proof}
The proof is similar to the proof of \autoref{cor:corollary 1 student}, the only difference lies in the fact that \eqref{prf:third term student} is replaced by 
\begin{align*}
\mathbb{E}_{(\hat{\rho}_{1},\dots,\hat{\rho}_{T})}\mathbb{E}_{\bc\sim \rho_{k}}\sum_{t=1}^{T}\frac{\lambda_{t-1}}{2}[\ell(\bc,x_{t})-\ell(\hat{\bc}_{t},x_{t})]^2
\leq C_{1}^{2}\sqrt{T \log T}.
\end{align*}
\end{proof}

%\section{Numerical experiments}\label{appendixB}

\begin{figure}[h!]
  \begin{subfigure}{\linewidth}
    \centering
    \includegraphics[width=0.49\linewidth, height=.16\textheight]{PACBO}\hfill
    \includegraphics[width=0.49\linewidth, height=.16\textheight]{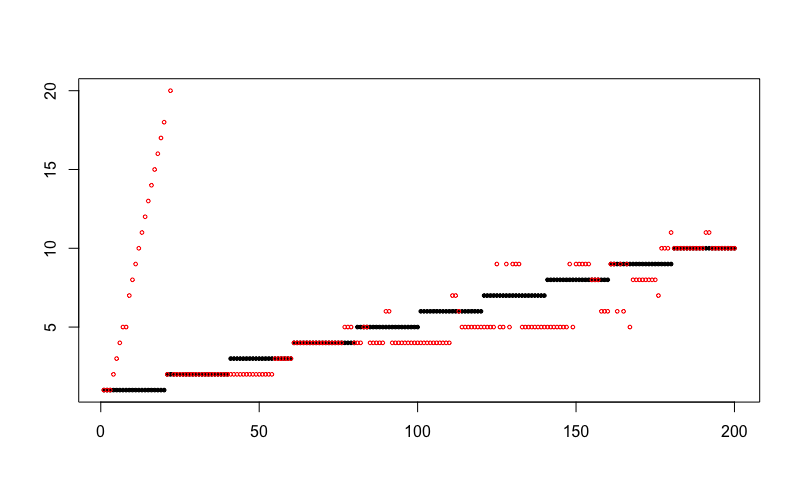}
    \caption{\name\ (left) and Silhouette (right)}
  \end{subfigure}
  \begin{subfigure}{\linewidth}
    \centering
    \includegraphics[width=0.49\linewidth, height=.16\textheight]{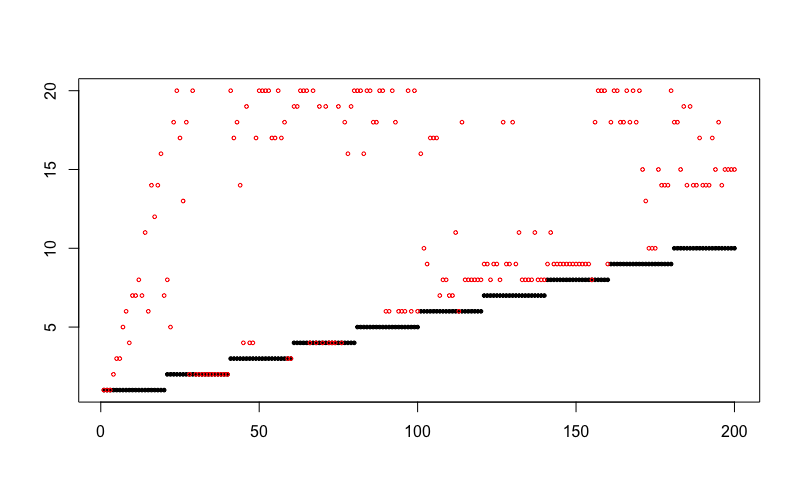}\hfill
    \includegraphics[width=0.49\linewidth, height=.16\textheight]{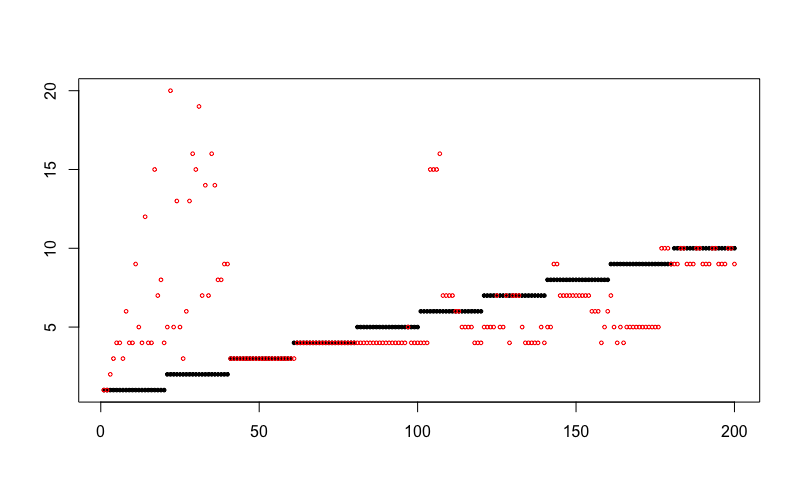}
        \caption{Calinski (left) and Hartigan (right)}
  \end{subfigure}
  \begin{subfigure}{\linewidth}
    \centering
    \includegraphics[width=0.49\linewidth, height=.16\textheight]{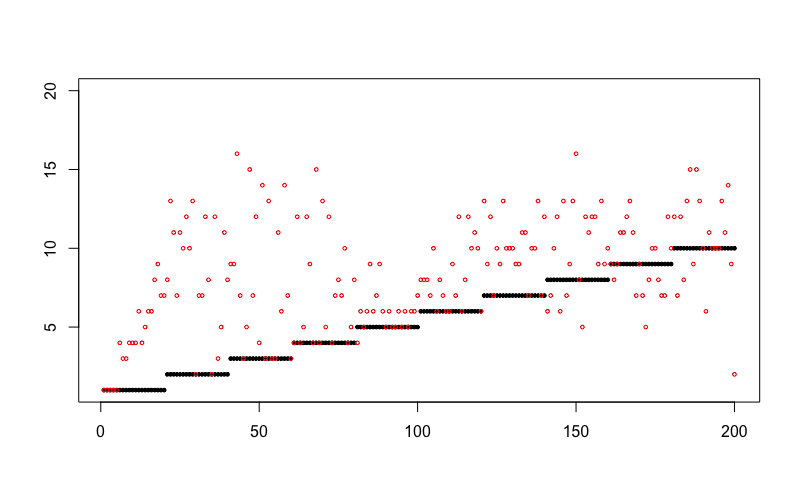}\hfill
    \includegraphics[width=0.49\linewidth, height=.16\textheight]{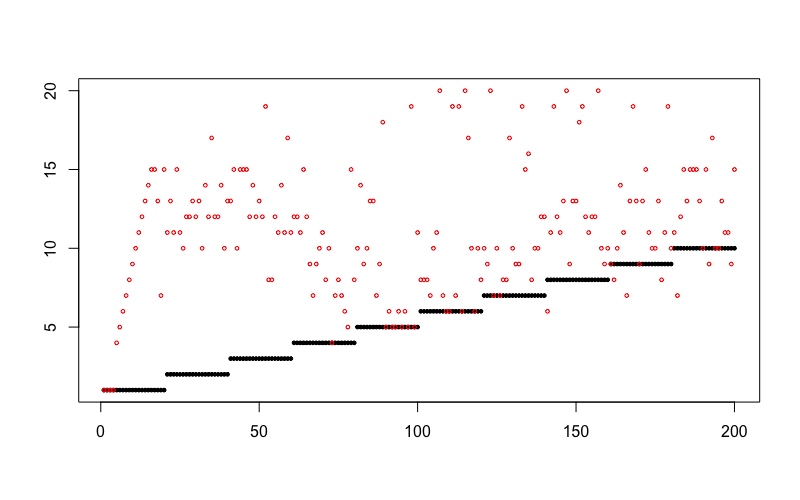}
        \caption{Djump (left) and DDSE (right)}
  \end{subfigure}
  \begin{subfigure}{\linewidth}
    \centering
    \includegraphics[width=0.49\linewidth, height=.16\textheight]{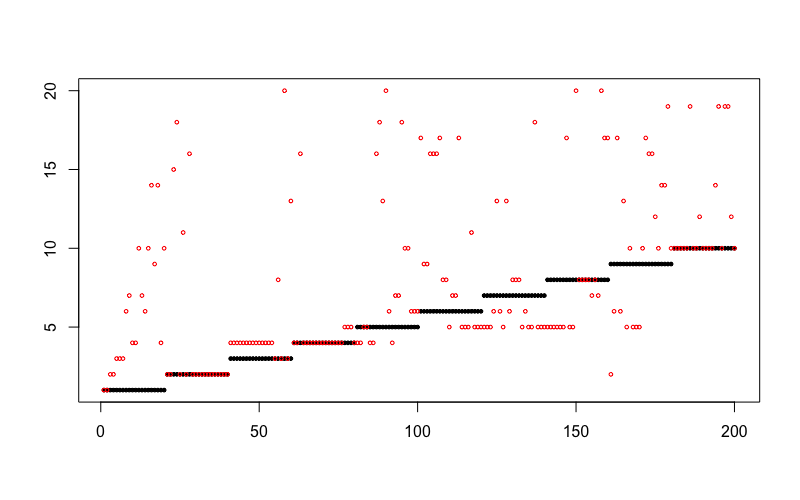}\hfill
    \includegraphics[width=0.49\linewidth, height=.16\textheight]{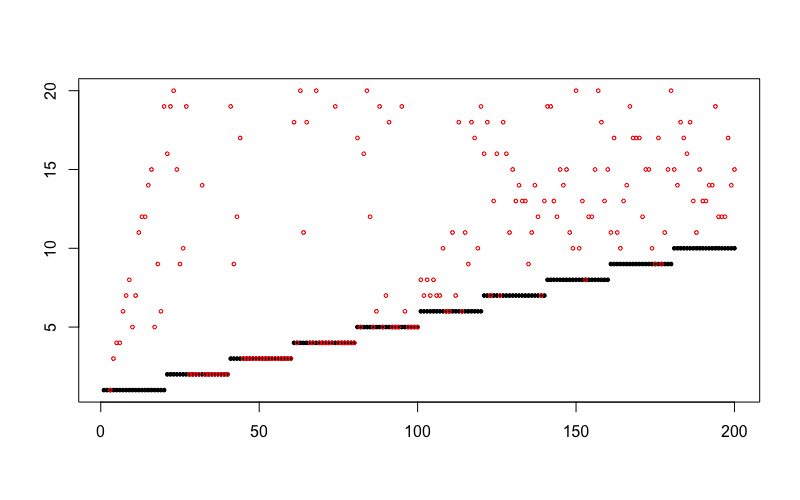}
        \caption{Lai (left) and Gap (right)}
  \end{subfigure}
      \caption{True (black) and estimated (red) number of clusters as functions of $t$.}
          \label{fig6}
\end{figure}

For the sake of completion, we present in \autoref{fig6} the performance of PACBO and its seven competitors for estimating the true number $k^\star_t$ of clusters along time. We acknowledge that no theoretical guarantee is derived for the estimation of $k^\star_t$ yet the practical behavior is remarkable.

\end{document}